\documentclass[preprint,12pt]{elsarticle}
\usepackage{graphicx} 
\usepackage{subfigure}
\usepackage{float}
\usepackage[boxed,linesnumbered]{algorithm2e}

\usepackage[inline,nomargin]{fixme}

\usepackage{pseudo}
\usepackage{array}
\usepackage{url}

\usepackage{color}

\usepackage{hyperref}
\usepackage{amsmath,amssymb,amsfonts}
\usepackage{bm,bbm}

\usepackage{multirow}

\DeclareMathOperator*{\argmin}{argmin}
\DeclareMathOperator*{\argmax}{argmax}

\def\beq{\begin{equation}}
\def\eeq{\end{equation}}

\journal{Artificial Intelligence}

\def \be{\begin{equation}}
\def \ee{\end{equation}}

\newtheorem{example}{Example}[section]

\def \argmax{\mbox{argmax}}

\newcommand{\ignore}[1]{}

\newcommand{\defas}{\ensuremath{\stackrel{\text{\tiny        def}}{=}}\xspace}
\newcommand{\imp}{\ensuremath{\rightarrow}\xspace}

\newcommand{\ignoreinshort}[1]{\textcolor{green}{#1}} 

\renewcommand{\ignoreinshort}[1]{}

\newcommand{\LL}{\textsc{LL}\xspace}
\newcommand{\optimathsat}{\textsc{OptiMathSAT}\xspace}
\newcommand{\mathsatfive}{\textsc{MathSAT5}\xspace}
\newcommand{\zthree}{\textsc{Z3}\xspace}
\newcommand{\yices}{\textsc{Yices}\xspace}
\newcommand{\cvcfour}{\textsc{CVC4}\xspace}

\newcommand{\T}{\ensuremath{\mathcal{T}}\xspace}

\newcommand{\tmany}{\ensuremath{\bigcup_i \T_i}\xspace}

\newcommand{\TsolversGen}[1]{\ensuremath{{#1}\textit{-solvers}}\xspace}

\newcommand{\Tsolvers}{\TsolversGen{\T}}

\newcommand{\euf}{\ensuremath{\mathcal{EUF}}\xspace}

\newcommand{\larat}{\ensuremath{\mathcal{LRA}\xspace}}
\newcommand{\laint}{\ensuremath{\mathcal{LIA}\xspace}}

\newcommand{\nla}{\ensuremath{\mathcal{NLA}}\xspace}
\newcommand{\nlarat}{\ensuremath{\mathcal{NLA}(\mathbb{R})}\xspace}

\renewcommand{\nlarat}{\ensuremath{\mathcal{NLRA}}\xspace}
 
\newcommand{\bv}{\ensuremath{\mathcal{BV}}\xspace}
\newcommand{\mem}{\ensuremath{\mathcal{AR}}\xspace}

\newcommand{\omlarat}{\ensuremath{\text{OMT}(\larat)}\xspace}
\newcommand{\omlaratplus}{\ensuremath{\text{OMT}(\larat\cup\T)}\xspace}

\begin{document}

\begin{frontmatter}



	\title{Structured Learning Modulo Theories}


	\author[label1]{Stefano Teso}
	\ead{steteso@fbk.eu}
	\author[label2]{Roberto Sebastiani}
	\ead{roberto.sebastiani@unitn.it}
	\author[label2]{Andrea Passerini}
	\ead{passerini@disi.unitn.it}

	\address[label1]{DKM - Data \& Knowledge Management Unit, \\ 
Fondazione Bruno Kessler, \\
Via Sommarive, 18, I38123 Povo, TN, Italy.}

	\address[label2]{DISI - Dipartimento di Ingegneria e Scienza dell'Informazione, \\
Universit\`a degli Studi di Trento, \\
Via Sommarive 5, I-38123 Povo, TN, Italy.}

	\begin{abstract}
\noindent Modelling problems containing a mixture of Boolean and numerical
variables is a long-standing interest of Artificial Intelligence. However,
performing inference and learning in hybrid domains is a particularly daunting
task. The ability to model this kind of domains is crucial in ``learning to
design'' tasks, that is, learning applications where the goal is to learn from
examples how to perform automatic {\em de novo} design of novel objects. In
this paper we present Structured Learning Modulo Theories, a max-margin
approach for learning in hybrid domains based on Satisfiability Modulo
Theories, which allows to combine Boolean reasoning and optimization over
continuous linear arithmetical constraints. The main idea is to leverage a
state-of-the-art generalized Satisfiability Modulo Theory solver for
implementing the inference and separation oracles of Structured Output SVMs. We
validate our method on artificial and real world scenarios.

	\end{abstract}


\end{frontmatter}



\section{Introduction}

Research in machine learning has progressively widened its scope, from
simple scalar classification and regression tasks to more complex
problems involving multiple related variables. Methods developed in
the related fields of statistical relational learning
(SRL)~\cite{getoor2007introduction} and structured-output
learning~\cite{BakHofSchSmoTasVis07} allow to perform learning, reason
and make inference about relational entities characterized by both
hard and soft constraints. Most methods rely on some form of (finite)
First-Order Logic (FOL) to encode the learning problem, and define the
constraints as weighted logical formulae. One issue with these
approaches is that First-Order Logic is not suited for efficiently
reasoning over hybrid domains, characterized by both continuous and
discrete variables.  The Booleanization of an $n$-bit integer variable
requires $2^n$ distinct Boolean states, making naive translation impractical;
for rational variables the situation is even worse.
 In addition,
standard FOL automated reasoning techniques offer no mechanism to deal
efficiently with operators among numerical variables, like comparisons
(e.g.  ``less-than'', ``equal-to'') and arithmetical operations (e.g.
summation), limiting the range of realistically applicable constraints
to those based solely on logical connectives. On the other hand, many
real-world domains are inherently hybrid and require to reason over
inter-related continuous and discrete variables. This is especially
true in \emph{constructive} machine learning tasks, where the focus is
on the \emph{de-novo} design of objects with certain characteristics
to be learned from examples (e.g. a recipe for a dish, with
ingredients, proportions, etc.).

There is relatively little previous work on {\em hybrid} SRL
methods. A number of approaches~\cite{Lip09,BroMihGet10,Kuz11,Dil12}
focused on the feature representation perspective, in order to extend
statistical relational learning algorithms to deal with continuous
features as {\em inputs}. On the other hand, performing inference over
joint continuous-discrete relational domains is still a challenge. The
few existing
attempts~\cite{church08,Wan08,NarBusKon10,Gut11,ChoAmi12,IslRamRam2012}
aim at extending statistical relational learning methods to the hybrid
domain. All these approaches focus on modeling the probabilistic
relationships between variables. While this allows to compute marginal
probabilities in addition to most probable configurations, it imposes
strong limitations on the type of constraints they can
handle. Inference is typically run by approximate methods, based on
variational approximations or sampling strategies.  
Exact inference, support for hard numeric (in addition to Boolean)
constraints and combination of diverse theories, like linear algebra
over rationals and integers, are out of the scope of these
approaches. Hybrid Markov Logic networks~\cite{Wan08} and
Church~\cite{church08} are the two formalisms which are closer to the
scope of this paper. Hybrid Markov Logic networks~\cite{Wan08} extend
Markov Logic by including continuous variables, and allow to embed
numerical comparison operators (namely $\ne$, $\geq$ and $\leq$) into
the constraints by defining an {\em ad hoc} translation of said
operators to a continuous form amenable to numerical optimization.
Inference relies on a stochastic local search procedure that
interleaves calls to a MAX-SAT solver and to a numerical optimization
procedure.  This inference procedure is incapable of dealing with hard
numeric constraints because of the lack of feedback from the
continuous optimizer to the satisfiability module.
Church~\cite{church08} is a very expressive
probabilistic programming language that can potentially represent
arbitrary constraints on both continuous and discrete variables. Its
focus is on modelling the generative process underlying the program,
and inference is based on sampling techniques. This makes inference
involving continuous optimization subtasks and hard constraints
prohibitively expensive, as will be discussed in the experimental
evaluation.

In order to overcome the limitations of existing approaches, we
focused on the most recent advances in automated reasoning over hybrid
domains. Researchers in automated reasoning and formal verification
have developed logical languages and reasoning tools that allow for
{\em natively} reasoning over mixtures of Boolean {\em and} numerical
variables (or even more complex structures). These languages are
grouped under the umbrella term of {\em Satisfiability Modulo
  Theories} (SMT)~\cite{BSST09HBSAT}.  Each such language corresponds
to a decidable fragment of First-Order Logic augmented with an
additional background theory $\mathcal{T}$. There are many such
background theories, including those of linear arithmetic over the
rationals \larat\ or over the integers \laint, among
others~\cite{BSST09HBSAT}. In SMT, a formula can contain Boolean
variables (i.e. 0-ary logical predicates) and connectives, mixed with
symbols defined by the theory $\mathcal{T}$, e.g. rational variables
and arithmetical operators. For instance, the SMT(\larat) syntax
allows to write formulas such as:
$$ 
{\tt touching\_i\_j}  \leftrightarrow ((x_i + dx_i = x_j) \lor (x_j +
dx_j = x_i))
$$
where the variables are Boolean (${\tt touching\_i\_j}$) or rational
($x_i$, $x_j$, $dx_i$,$dx_j$).%
\footnote{Note that SMT solvers handle also formulas on {\em combinations}
  of theories, like e.g.\\
$
(d\ge 0)\wedge (d<1) \wedge 
(   
    (f(d) = f(0)) \imp ({\tt read}({\tt write}(V,i,x),i+d)=x+1)  )     
$, 
where $d,\ i,\ x$ are integer
variables, $V$ is an array variable, 
$f$ is an uninterpreted function symbol,
 {\tt read} and {\tt write} are functions of the theory
 of arrays \cite{BSST09HBSAT}.
However, for the scope of this paper it suffices to consider the
\larat{} and \laint{} theories, and their combination. 
}
More specifically, SMT is a decision
problem, which consists in finding an assignment to the variables of a
quantifier-free formula, both Boolean and theory-specific ones, that
makes the formula true, and it can be seen as an extension of SAT.

Recently, researchers have leveraged SMT from decision to
optimization.
In particular, {\em MAX-SAT Modulo Theories} ({\em MAX-SMT})
\cite{nieuwenhuis_sat06,cimattifgss10,cgss_sat13_maxsmt}
generalizes MAX-SAT \cite{LM09HBSAT} to SMT
formulae, and consists in finding a theory-consistent truth
assignment to the atoms of the input SMT formula $\varphi$ which maximizes the
total weight of the satisfied clauses of $\varphi$.
More generally, 
{\em Optimization
Modulo Theories} ({\em OMT})~\cite{nieuwenhuis_sat06,st-ijcar12,sebastiani14_optimathsat,li_popl14}
consists in finding  
a {\em model} for  $\varphi$ which minimizes the value of some
(arithmetical) term, and strictly subsumes MAX-SMT
\cite{sebastiani14_optimathsat}. 
Most important for the scope of this paper is that there are
high-quality 
OMT solvers which, at least for the $\larat$ theory, can handle
problems with thousands of hybrid variables.

In this paper we propose {\em Learning Modulo Theories} (LMT), a class
of novel hybrid statistical relational learning methods. The main idea
is to combine a solver able to deal with Boolean and rational
variables with a structured output learning method. In particular, we
rely on structured-output Support Vector Machines
(SVM)~\cite{tsochantaridis2005large,joachims2009predicting}, a very
flexible max-margin structured prediction method. Structured-output
SVMs are a generalization of binary SVM classifiers to predict
structured outputs, like sequences or trees. They generalize the
max-margin principle by learning to separate correct from incorrect
output structures with a large margin. Training structured-output SVMs
requires a separation oracle for generating counter-examples and
update the parameters, while using them at prediction stage requires
an inference oracle generating the highest scoring candidate structure
for a certain input. In order to implement the two oracles, we
leverage a state-of-the-art OMT solver.  This combination enables LMT
to perform learning and inference in mixed Boolean-numerical domains.
Thanks to the efficiency of the underlying OMT solver, and of the
cutting plane algorithm we employ for weight learning, LMT is capable
of addressing constructive learning problems which cannot be
efficiently tackled with existing methods. Furthermore, LMT is {\em
  generic}, and can in principle be applied to any of the existing
background theories. This paper builds on a previous work in which
MAX-SMT was used for interactive preference
elicitation~\cite{stochLogicUtFun2010}. Here we focus on generating
novel structures/configurations from few prototypical examples, and
cast the problem as supervised structured-output
learning. Furthermore, we increase the expressive power from MAX-SMT
to full OMT. This allows to model much richer cost functions, for
instance by penalizing an unsatisfied constraint by a cost
proportional to the distance from satisfaction.


The rest of the paper is organized as follows. In
Section~\ref{sec:related} we review the relevant related work, with an
in-depth discussion on all hybrid approaches and their relationships
with our proposed framework. Section~\ref{sec:smt} provides an
introduction to SMT 
and OMT technology. Section~\ref{sec:method} reviews structured-output
SVMs and shows how to cast LMT in this learning
framework. Section~\ref{sec:apps} reports an experimental evaluation
showing the potential of the approach. Finally, conclusions are drawn
in Section~\ref{sec:conc}.

\section{Related Work} 
\label{sec:related}

There is a body of work concerning integration of relational and
numerical data from a feature representation perspective, in order to
effectively incorporate numerical features into statistical relational
learning models. Lippi and Frasconi~\cite{Lip09} incorporate neural
networks as feature generators within Markov Logic Networks, where
neural networks act as numerical functions complementing the Boolean
formulae of standard MLNs. Semantic Based Regularization~\cite{Dil12}
is a framework for integrating logic constraints within kernel
machines, by turning them into real-valued constraints using
appropriate transformations (T-norms). The resulting optimization
problem is no longer convex in general, and they suggest a stepwise
approach adding constraints in an incremental fashion, in order to
solve progressively more complex problems. In Probabilistic Soft
Logic~\cite{BroMihGet10}, arbitrarily complex similarity measures
between objects are combined with logic constraints, again using
T-norms for the continuous relaxation of Boolean operators.  In
Gaussian Logic~\cite{Kuz11}, numeric variables are modeled with
multivariate Gaussian distributions. Their parameters are tied
according to logic formulae defined over these variables, and combined
with weighted first order formulae modeling the discrete part of the
domain (as in standard MLNs). All these approaches aim at extending
statistical relational learning algorithms to deal with continuous
features as {\em inputs}. On the other hand, our framework aims at
allowing learning and inference over hybrid continuous-discrete
domains, where continuous and discrete variables are the {\em output}
of the inference process.

While a number of efficient lifted-inference algorithms have been
developed for Relational Continuous
Models~\cite{ChoHilAmi10,ChoGuzAmi11,AhmKerSan11}, performing
inference over joint continuous-discrete relational domains is still a
challenge. The few existing attempts aim at extending statistical
relational learning methods to the hybrid domain.

Hybrid Probabilistic Relational Models~\cite{NarBusKon10} extend
Probabilistic Relational Models (PRM) to deal with hybrid domains by
specifying templates for hybrid distributions as standard PRM specify
templates for discrete distributions. A template instantiation over a
database defines a Hybrid Bayesian
Network~\cite{Murphy98hybrid,HybridBN}. Inference in Hybrid BN is
known to be hard, and restrictions are typically imposed on the
allowed relational structure (e.g. in conditional Gaussian models,
discrete nodes cannot have continuous parents). On the other hand, LMT
can accomodate arbitrary combinations of predicates from the theories
for which a solver is available. These currently include linear
arithmetic over both rationals and integers as well as a number of
other theories like strings, arrays and bit-vectors.

Relational Hybrid Models~\cite{ChoAmi12} (RHM) extend Relational
Continuous Models to represent combinations of discrete and continuous
distributions. The authors present a family of lifted variational
algorithms for performing efficient inference, showing substantial
improvements over their ground counterparts. As for most hybrid SRL
approaches which will be discussed further on, the authors focus on
efficiently computing probabilities rather than efficiently finding
optimal configurations. Exact inference, hard constraints and theories
like algebra over integers, which are naturally handled by our LMT
framework, are all out of the scope of these approaches.  Nonetheless,
lifted inference is a powerful strategy to scale up inference and
equipping OMT and SMT tools with lifting capabilities is a promising
direction for future improvements.

The PRISM~\cite{Sato95} system provides primitives for Gaussian
distributions. However, inference is based on proof enumeration, which
makes support for continuous variables very limited. Islam et
al.~\cite{IslRamRam2012} recently extended PRISM to perform inference
over continuous random variables by a {\it symbolic} procedure which
avoids the enumeration of individual proofs. The extension allows to
encode models like Hybrid Bayesian Networks and Kalman Filters. Being
built on top of the PRISM system, the approach assumes the exclusive
explanation and independence property: no two different proofs for the
same goal can be true simultaneously, and all random processes within
a proof are independent (some research directions for lifting these
restrictions have been suggested~\cite{IslamPhd12}). LMT has no
assumptions on the relationships between proofs.

Hybrid Markov Logic Networks~\cite{Wan08} extend Markov Logic Networks
to deal with numeric variables. A Hybrid Markov Logic Network consists
of both First Order Logic formulae and numeric terms. Most probable
explanation (MPE) inference is performed by a hybrid version of
MAXWalkSAT, where optimization of numeric variables is performed by a
general-purpose global optimization algorithm (L-BFGS). This approach
is extremely flexible and allows to encode arbitrary numeric
constraints, like soft equalities and inequalities with quadratic or
exponential costs. A major drawback of this flexibility is the
computational cost, as each single inference step on continuous
variables requires to solve a global optimization problem, making the
approach infeasible for addressing medium to large scale problems.
Furthermore, this inference procedure is incapable of dealing with
hard constraints involving numeric variables, as can be found for
instance in layout problems (see e.g. the constraints on touching
blocks or connected segments in the experimental evaluation). This is
due to the lack of feedback from the continuous optimizer to the
satisfiability module, which should inform about conflicting
constraints and help guiding the search towards a more promising
portion of the search space. Conversely, the OMT technology underlying
LMT is built on top of SMT solvers and is hence specifically designed
to tightly integrate theory-specific and SAT
solvers~\cite{nieuwenhuis_sat06,cimattifgss10,st-ijcar12,sebastiani14_optimathsat,li_popl14}.
Note that the tight interaction between theory-specific and modern
CDCL SAT solvers, plus many techniques developed for maximizing their
synergy, are widely recognised as one key reason of the success of SMT
solvers \cite{BSST09HBSAT}.  Note also that previous attempts to
substitute standard SAT solvers with WalkSAT inside an SMT solver have
failed, producing dramatic worsening of performance~\cite{frocos11_walksmt}.  

Hybrid ProbLog~\cite{Gut11} is an extension of the probabilistic logic
language ProbLog~\cite{problog07} to deal with continuous variables. A
ProbLog program consists of a set of probabilistic Boolean facts, and
a set of deterministic first order logic formulae representing the
background knowledge. Hybrid ProbLog introduces a set of probabilistic
continuous facts, containing both discrete and continuous
variables. Each continuous variable is associated with a probability
density function. The authors show how to compute the probability of
success of a query, by partitioning the continuous space into
admissible intervals, within which values are interchangeable with
respect to the provability of the query. The drawback of this approach
is that in order to make this computation feasible, severe limitations
have to be imposed on the use of continuous variables. No algebraic
operations or comparisons are allowed between continuous variables,
which should remain uncoupled. Some of these limitations have been
overcome in a recent approach~\cite{GutThoKim11} which performs
inference by forward (i.e. from facts to rules) rather than backward
reasoning, which is the typical inference process in (probabilistic)
logic programming engines (SLD-resolution and its probabilistic
extensions). Forward reasoning is more amenable to be adapted to
sampling strategies for performing approximate inference and dealing
with continuous variables. On the other hand, inference by sampling
makes it prohibitively expensive to reason with hard continuous
constraints.

Church~\cite{church08} is a very expressive probabilistic programming
language that can easily accomodate hybrid discrete-continuous
distributions and arbitrary constraints. In order to deal with the
resulting complexity, inference is again performed by sampling
techniques, which result in the same up-mentioned limitations. Indeed,
our experimental evaluation shows that Church is incapable of solving
in reasonable time the simple task of generating a pair of blocks
conditioned on the fact that they touch somewhere.~\footnote{The only
  publicly available version of Hybrid ProbLog is the original one by
  Gutmann, Jaeger and De Raedt~\cite{Gut11} which does not support
  arithmetic over continuous variables. However we have no reason to
  expect the more recent version based on sampling should have a
  substantially different behaviour with respect to what we observe
  with Church.}

An advantage of these probabilistic inference approaches is that they
allow to return marginal probabilities in addition to most probable
explanations. This is actually the main focus of these approaches, and
the reason why they are less suitable for solving the latter problem
when the search space becomes strongly disconnected. As most
structured-output approaches over which it builds, LMT is currently
limited to the task of finding an optimal configuration, which in a
probabilistic setting corresponds to generating the most probable
explanation. We are planning to extend it to also perform probability
computation, as discussed in the conclusions of the paper.

\section{From Satisfiability to Optimization Modulo Theories} 
\label{sec:smt}

Propositional satisfiability (SAT), is the problem of deciding whether a
logical formula over Boolean variables and logical connectives can be satisfied
by some truth value assignment of the Boolean variables.~%
\footnote{CDCL SAT-solving algorithms, and SMT-solving ones thereof,
 require the input formula to be in conjunctive normal form (CNF), 
i.e., a conjunction of {\em clauses}, each clause being a
disjunction of propositions or of their negations.
Since they very-effectively pre-convert input formulae into CNF
 \cite{Pre09HBSAT}, we assume  wlog   input formulae to have any
 form. 
}
%
In the last two decades we have witnessed an impressive advance
in the efficiency of SAT solvers, which nowadays can handle
industrial-derived formulae in the order of up to $10^6-10^7$
variables. 
%
Modern SAT solvers are based on the conflict-driven clause-learning
(CDCL) schema \cite{MSLM09HBSAT}, and adopt a variety of 
very-efficient search techniques \cite{HandbookOfSAT2009}.

\ignoreinshort{
MAX-SAT \cite{LM09HBSAT} leverages SAT from decision to optimization.
In a (weighted partial) MAX-SAT problem a subset of the clauses 
of the input formula $\varphi$ ---called ``soft'' clauses---
 are given a positive numerical weight, and the problem 
consists in finding a truth assignment to the atoms of $\varphi$ 
which satisfies all the other clauses ---called ``hard'' clauses--- and 
maximizes the total weight of the satisfied soft clauses.
A wide variety of MAX-SAT tools have been proposed in the literature
and are currently available.
}

%
In the contexts of automated reasoning (AR) and formal verification (FV), 
important {\em decision} problems are effectively encoded into and solved as
Satisfiability Modulo Theories (SMT) problems \cite{MouraB11}.
SMT is the problem of deciding the
satisfiability of a (typically quantifier-free) first-order formula
with respect to some decidable {\em background theory} \T,
which can also be a combination of theories \tmany. 
%
Theories of practical interest are, e.g., those of 
equality and uninterpreted functions (\euf), of linear arithmetic over the
rationals (\larat{}) or over the integers (\laint{}), 
of non-linear arithmetic over the reals (\nla), of arrays (\mem),
of bit-vectors (\bv), and their combinations.

\ignoreinshort{%
\begin{example}
Consider the following toy $\laint\cup\euf\cup\mem$-formula:
$$
\varphi\defas  (d\ge 0)\wedge (d<1) \wedge 
(   
    (f(d) = f(0)) \imp (read(write(V,i,x),i+d)=x+1)  )     
$$
where  $d,\ i,\ x$ are integer
variables, $V$ is an array variable, 
$f$ is an uninterpreted function symbol,
 $read(),\ write()$ are the standard functions of \mem theory.~%
\footnote{$read(V,i)$ returns the value of the $i$-th element of
  array $V$, and $write(V,i,x)$ returns a new array resulting from
  writing value $x$ in the $i$-th element of
  array $V$. 
}
$\varphi$ is inconsistent in the combined theory
$\laint\cup\euf\cup\mem$.
In fact, from the first two atoms $(d\ge 0)$  $(d<1)$ we deduce in \laint that $d=0$, from
which we  deduce in \euf that the second atom $(f(d) = f(0))$ must
be true; hence, by Boolean reasoning, we deduce that the last atom 
$(read(write(V,i,x),i+d)=x+1)$ must be true; in \laint we 
deduce $ (read(write(V,i,x),i)=x+1)$ and hence 
$\neg (read(write(V,i,x),i)=x)$. The  latter contradicts one of
the axioms in  \mem,~%
\footnote{``$\forall\ V,i,x.\ (read(write(V,i,x),i)=x)$'' is an axiom of
  \mem{}, meaning ``the value you read in the $i$-th element of the array resulting from
  writing $x$ into the $i$-th element in the array $V$, is $x$.}
so that we can deduce that $\varphi$ is unsatisfiable in $\laint\cup\euf\cup\mem$.
\end{example}
}
%
In the last decade efficient SMT
solvers have been developed following the so-called {\em lazy
approach}, that combines the power of modern 
CDCL  SAT solvers 
with the expressivity of dedicated decision procedures (\Tsolvers)
for several first-order theories of interest.
Modern lazy SMT solvers ---like e.g. \cvcfour
\footnote{\url{http://cvc4.cs.nyu.edu/}}, \mathsatfive
\footnote{\url{http://mathsat.fbk.eu/}}, \yices
\footnote{\url{http://yices.csl.sri.com/}}, \zthree
\footnote{\url{ttp://research.microsoft.com/en-us/um/redmond/projects/z3/ml/z3.html}}---
combine a variety of solving techniques coming
from very hetherogeneous domains\ignoreinshort{, ranging from SAT (e.g. for Boolean
reasoning and the \bv theory), first-order theorem proving (e.g., for \euf
and \mem) operations research and CSP (e.g., for \larat, \laint or
\nlarat), and computer algebra (e.g., for \nlarat)}.  We refer the reader
to \cite{sebastiani07,BSST09HBSAT} for an overview on lazy SMT
solving, and to the URLs of the above solvers for a description of
their supported theories and functionalities.

More recently, 
SMT has also been leveraged from decision to optimization.
\ignoreinshort{
First, {\em MAX-SAT Modulo Theories} ({\em MAX-SMT})
generalizes MAX-SAT  to SMT
formulae, by finding a {\em theory-consistent} truth
assignment to the atoms of the input SMT formula $\varphi$ which 
maximizes the total weight of the satisfied soft clauses.
A few MAX-SMT tools have been proposed in the literature
and are currently available
\cite{nieuwenhuis_sat06,cimattifgss10,cgss_sat13_maxsmt},
which are built on top of state-of-the-art SMT solvers.~%
\footnote{Also \yices and \zthree provide support for
  MAX-SMT, but we are not aware of any public document describing the
  procedures used there.}
}
\ignoreinshort{
Second,  and more generally, 
}
{\em Optimization
Modulo Theories} ({\em OMT})~\cite{nieuwenhuis_sat06,st-ijcar12,sebastiani14_optimathsat,li_popl14},
is the problem of finding a {\em model} for an SMT formula 
 $\varphi$ which minimizes the value of some
arithmetical cost function. 
References \cite{st-ijcar12,sebastiani14_optimathsat,li_popl14} present some general OMT procedure 
adding to SMT the capability of finding
models minimizing cost functions in \larat. 
This problem is denoted \omlarat{} if only the \larat{} theory is
involved in the SMT formula, \omlaratplus{} if some other 
theories are involved. 
\ignoreinshort{~%
\footnote{
In \cite{st-ijcar12,sebastiani14_optimathsat}, \larat{} and \T are combinations of Nelson-Oppen
theories, as described in~\cite{bozzanobcjrrs06}.  
(e.g., \euf, \larat, and \mem match this definition.)}
}
Such procedures combine standard lazy SMT-solving 
with LP minimization techniques\ignoreinshort{, implementing mixed
binary-rearch/branch\&bound minimization strategies inside the CDCL
schema of the underlining SAT solver, so as to fully exploit the
pruning power of the CDCL paradigm}. \omlaratplus procedures have been
implemented into the \optimathsat  tool,\footnote{
\url{http://optimathsat.disi.unitn.it/}}
 a sub-branch of \mathsatfive{}. 
\ignoreinshort{%
Importantly, \omlarat and \omlaratplus are strictly more expressive
than MAX-SMT, since the latter can be straightforwardly and
effectively encoded into the former, but not vice
versa~\cite{st-ijcar12,sebastiani14_optimathsat}.~%
\footnote{The effectiveness of such encoding is witnessed
  by the 
  experimental evaluation of \cite{cgss_sat13_maxsmt}, where the specialized 
  MAX-SMT tool \LL introduced there did not beat \optimathsat on MAX-SMT
  benchmarks, despite both tools being built on top of \mathsatfive.}
}

\begin{example}
 Consider the following toy \larat-formula $\varphi$:
$$
(cost=x+y)\wedge
(x\ge 0) \wedge
(y\ge 0) \wedge
(A \vee (4x+y-4\ge 0)) \wedge
(\neg A \vee (2x+3y-6\ge 0))
$$
and the \omlarat problem of finding the model of $\varphi$ (if any) 
which makes the value of $cost$ minimum. 
In fact, depending on the truth value of $A$,
there are two possible alternative sets of constraints to minimize:
$$
\begin{array}{rll}
\{A,(cost=x+y),(x\ge 0),(y\ge 0),(2x+3y-6\ge 0)\} & \\
\{\neg A,(cost=x+y),(x\ge 0),(y\ge 0),(4x+y-4\ge 0)\} & 
\end{array}
$$
whose minimum-cost models are, respectively:
$$
\begin{array}{lll}
 \{A=True,x=0.0,y=2.0,cost=2.0\} \\
 \{A=False,x=1.0,y=0.0,cost=1.0\} 
\end{array}
$$
from which we can conclude that the latter is a minimum-cost model for $\varphi$.
\end{example}

Overall, for the scope of this paper, it is important to highlight the
fact that OMT solvers are available which, thanks to the underlying
SAT and SMT technologies, can handle problems with a large number of
hybrid variables (in the order of thousands, at least for the $\larat$
theory).

To this extent, we notice that the underlying theories and \Tsolvers
provide the meaning and the reasoning capabilities for specific
predicates and function symbols (e.g., the \larat-specific symbols 
``$\ge$" and ``$+$'', or the \mem-specific symbols ``{\tt
  read(...)}'', ``{\tt write(...)}'') that would otherwise be very
difficult to describe, or to reason over, with logic-based automated
reasoning tools ---e.g., traditional first-order theorem provers
cannot handle arithmetical reasoning efficiently--- or with
arithmetical ones ---e.g., DLP, ILP, MILP, LGDP tools
\cite{balasdisjunctive,lodi,sawayagrossmann2012} or CLP tools
\cite{Jaffar94,CodDia96,JafJox92} do not handle {\em symbolic}
theory-reasoning on theories like \euf{} or \mem.  Also, the
underlying CDCL SAT solver allows SMT solvers to handle a large amount of
Boolean reasoning very efficiently, which is typically out
of the reach of both first-order theorem provers and arithmetical
tools.

These facts
 motivate our choice of using SMT/OMT technology, and
hence the tool \optimathsat, as workhorse engines for reasoning in
hybrid domains.
Hereafter in the paper we consider only plain \omlarat{}.

Another prospective advantage of SMT technology is that modern SMT
solvers (e.g., \mathsatfive, \zthree, \ldots) 
have an {\em incremental
interface}, which allows 
for solving sequences of
``similar'' formulae without restarting the search from scratch at
each new formula, and instead reusing ``common'' parts of the
search performed for previous formulae (see, e.g.,
\cite{mathsat5_tacas13}). 
This drastically improves overall
performance on sequences of similar formulae. 
An incremental extension of \optimathsat{}, fully exploiting that of
\mathsatfive, is currently available.

Note that a current limitation of SMT solvers is that, unlike
traditional theorem provers, they typically handle efficiently only
quantifier-free formulae. Attempts at extending SMT to quantified
formulae have been made in the
literature~\cite{Rum08,BauTin11,Kruglov13}, and few SMT solvers (e.g.,
\zthree) do provide some support for quantified formulae. However, the
state of the art of these estensions is still far from being
satisfactory.  Nonetheless, the method we present in the paper can be
easily adapted to deal with this type of extensions once they will
reach the required level of maturity.

\section{Learning Modulo Theories using Cutting Planes}
\label{sec:method}

\newcommand{\lmtvar}[1]{\mbox{\boldmath{$#1$}}}
\newcommand{\lmtx}{\lmtvar{I}}
\newcommand{\lmty}{\lmtvar{O}}
\newcommand{\lmtw}{\lmtvar{w}}
\newcommand{\lmtphi}{\ensuremath{\varphi}}
\newcommand{\lmtPsi}{\lmtvar{\psi}}
\newcommand{\lmtxi}{\lmtvar{\xi}}
\newcommand{\lmtind}{\mathbbm{1}}

\subsection{An introductory example}

\begin{figure}[H]
	\centering
	\includegraphics{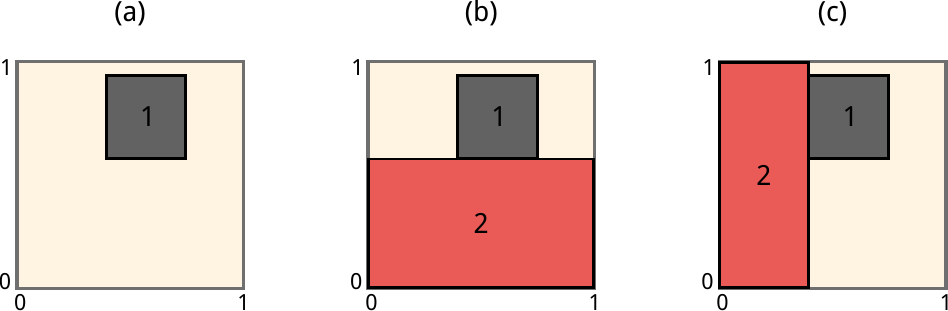}
	\caption{\label{fig:bwexample} (a) Initial configuration. (b) Optimal configuration
          for $w_1\ll w_2$. (c) Optimal configuration for $w_1\gg w_2$.}
\end{figure}

In order to introduce the LMT framework, we start with a toy learning example.
We are given a unit-length bounding box, $[0,1]\times[0,1]$, that contains a
given, fixed block (rectangle), as in Figure~\ref{fig:bwexample}~(a).
The block is identified by the four constants $(x_1,y_1,dx_1,dy_1)$, where
$x_1,y_1$ indicate the bottom-left corner of the rectangle, and $dx_1$,
$dy_1$ its width and height, respectively.
Now, suppose that we are assigned the task of fitting another block, identified
by the variables $(x_2,y_2,dx_2,dy_2)$, in the same bounding box, so as to
minimize the following {\em cost} function:
\beq
	cost := w_1 \times dx_2 + w_2 \times dy_2
	\label{eq:bwcost}
\eeq
with the additional requirements that (i) the two blocks ``touch'' either from
above, below, or sideways, and (ii) the two blocks do not overlap.

It is easy to see that the weights $w_1$ and $w_2$ control the shape and location of the
optimal solution.  If both weights are positive, then the cost is minimized by
any block of null size located along the perimeter of block 1. If both weights
are negative and $w_1\ll w_2$, then the optimal block will be placed so as to
occupy as much horizontal space as possible, while if $w_1\gg w_2$ it will
prefer to occupy as much vertical space as possible, as in
Figure~\ref{fig:bwexample}~(b,c). If $w_1$ and $w_2$ are close, then the
optimal solution depends on the relative amount of available vertical and
horizontal space in the bounding box. 

This toy example illustrates two key points.
First, the problem involves a mixture of {\em numerical variables} (coordinates, sizes of
block 2) and {\em Boolean variables}, along with {\em hard rules} that control the feasible space of
the optimization procedure (conditions (i) and (ii)), and {\em costs} --- or soft
rules --- which control the shape of the optimization landscape. This is the
kind of problem that can be solved in terms of Optimization Modulo Linear
Arithmetic, OMT(\larat).
Second, it is possible to estimate the weights $w_1$, $w_2$ from data in order
to learn what kind of blocks are to be considered optimal. In the following we
will describe how such a learning task can be framed within the structured output SVMs framework.

\subsection{Notation}

\begin{table}[H]
	\begin{tabular}{ll}
		\hline \hline
		{\bf Symbol}						& {\bf Meaning}			\\
		\hline
		{\tt above}, {\tt right}, \ldots			& Boolean variables		\\
		$x, y, dx, \ldots$					& Rational variables		\\
		(\lmtx, \lmty)						& Complete object; $\lmtx$ is the input, $\lmty$ is the output	\\
		$\lmtphi_1, \ldots, \lmtphi_m$				& Constraints			\\
		$\lmtind_k(\lmtx,\lmty)$				& Indicator for Boolean constraint $\lmtphi_k$ over $(\lmtx,\lmty)$	\\
		$c_k(\lmtx,\lmty)$					& Cost for arithmetical constraint $\lmtphi_k$ over $(\lmtx,\lmty)$	\\
		${\bf \lmtPsi}(\lmtx,\lmty)$				& Feature representation of the complete object \\
		$\psi_k(\lmtx,\lmty) := \lmtind_k(\lmtx,\lmty)$		& Feature associated to Boolean constraint $\lmtphi_k$		\\
		$\psi_k(\lmtx,\lmty) := -c_k(\lmtx,\lmty)$		& Feature associated to arithmetical constraint $\lmtphi_k$		\\
		\lmtw							& Weights			\\
		\hline \hline
	\end{tabular}
	\caption{\label{tab:notation} Explanation of the notation used throughout the text.}
\end{table}

We consider the problem of learning from a training set of $n$ complex objects
$\{ (\lmtx_i,\lmty_i) \}_{i=1}^n$, where each object $(\lmtx,\lmty)$ is represented as a set of
Boolean and rational variables:
$$ (\lmtx, \lmty) \in \underbrace{\left(\{\top,\bot\}\times\ldots\times\{\top,\bot\}\right)}_{\text{Boolean part}} \times \underbrace{\left(\mathbb{Q}\times\ldots\times\mathbb{Q}\right)}_{\text{rational part}} $$
We indicate Boolean variables using predicates such as ${\tt
  touching}(i,j)$, and write rational variables as lower-case letters,
e.g. $cost$, $distance$, $x$, $y$. Please note that while we write Boolean
variables using a First-Order syntax for readability, our method does require
the grounding of all Boolean predicates prior to learning and inference.
In the present formulation, we assume objects to be composed of two
parts: $\lmtx$ is the {\em
  input} (or observed) part, while $\lmty$ is the {\em output} (or
query) part.\footnote{We depart from the conventional $x$/$y$ notation for indicating
input/output pairs to avoid name clashes with the block coordinate variables.}
The learning problem is defined by a set of $m$ {\em constraints} $\{ \lmtphi_k
\}_{k=1}^m$. Each constraint $\varphi_k$ is either a Boolean- or
rational-valued function of the object $(\lmtx,\lmty)$.
For each Boolean-valued constraint $\varphi_k$, we denote its {\em indicator
function} as $\lmtind_k(\lmtx,\lmty)$, which evaluates to $1$ if the constraint is
satisfied and to $-1$ otherwise (the choice of $-1$ to represent falsity is
customary in the max-margin literature). Similarly, we refer to the {\em cost}
of a real-valued constraint $\varphi_k$ as $c_k(\lmtx,\lmty)\in\mathbb{Q}$.
The feature space representation of an object $(\lmtx,\lmty)$ is given
by the {\em feature vector} $\lmtPsi(\lmtx,\lmty)$, which is a
function of the constraints.  Each {\em soft} constraint $\varphi_k$
has an associated finite weight $w_k\in\mathbb{Q}$ (to be learned from
the data), while {\em hard} constraints have no associated weight. We
denote the vector of learned weights as
$\lmtw := \left( w_1, w_2, \ldots, w_m \right)$, and its Euclidean
norm as $\|\lmtw\|$. Table~\ref{tab:notation} summarizes the notation
used throughout the text.

\subsection{A Structural SVM approach to LMT}

Structured-output SVMs~\cite{tsochantaridis2005large} are a very flexible
framework that generalizes max-margin methods to the prediction of complex
outputs such as strings, trees and graphs. In this setting the association between
inputs $\lmtx$ and outputs $\lmty$ is controlled by a so-called {\em
compatibility function}
$f(\lmtx,\lmty) := \lmtw^\top \lmtPsi(\lmtx,\lmty)$
defined as a linear combination of the joint feature space representation
$\lmtPsi(\lmtx,\lmty)$ of the input-output pair.
Inference amounts to finding the most compatible output $\lmty^*$ for a given
input $\lmtx$, which equates to solving the following optimization problem:
\beq
	\lmty^* = \argmax_{\lmty} f(\lmtx,\lmty) = \argmax_{\lmty} \lmtw^\top \lmtPsi(\lmtx,\lmty)
	\label{eq:inference}
\eeq
Performing inference on structured domains is non-trivial, since the
maximization ranges over an exponential (and possibly unbounded) number of
candidate outputs.

Learning is formulated within the regularized empirical risk minimization
framework.  In order to learn the weights from a training set of $n$ examples,
one needs to define a non-negative {\em loss function}
$\Delta(\lmtx,\lmty,\lmty')$
that, for any given observation $\lmtx$, quantifies the penalty incurred when
predicting $\lmty'$ instead of the correct output $\lmty$. Learning can be
expressed as the problem of finding the weights $\lmtw$ that minimize the
per-instance error $\xi_i$ and the model complexity~\cite{tsochantaridis2005large}:
\beq
	\argmin_{\lmtw,\lmtxi} \; \frac{1}{2}\| {\lmtw} \|^2 + \frac{C}{n}\sum_{i=1}^n \xi_i
	\label{eq:learning}
\eeq
$$	\text{s.t.} \;\; \lmtw^\top (\lmtPsi(\lmtx_i,\lmty_i) - \lmtPsi(\lmtx,\lmty')) \geq \Delta(\lmtx_i,\lmty_i,\lmty') - \xi_i \;\; \forall \; i=1,\ldots,n;\;\lmty' \neq \lmty_i $$
Here the constraints require that the compatibility between any input
$\lmtx_i$ and its corresponding correct output $\lmty_i$ is always
higher than that with all wrong outputs $\lmty'$ by a margin, with
$\xi_i$ playing the role of per-instance violations. This formulation
is called $n$-slack margin rescaling and it is the original and most
accessible formulation of structured-output
SVMs. See~\cite{joachims2009cutting} for an extensive exposition of
alternative formulations.

Weight learning is a quadratic program, and can be solved very
efficiently with a cutting-plane (CP)
algorithm~\cite{tsochantaridis2005large}. Since in
Eq~(\ref{eq:learning}) there is an exponential number of constraints,
it is infeasible to naively account for all of them during
learning. Based on the observations that the constraints obey a
subsumption relation, the CP algorithm~\cite{joachims2009cutting}
sidesteps the issue by keeping a working set of active constraints
$\mathcal{W}$: at each iteration, it augments the working set with the
most violated constraint, and then solves the corresponding reduced
quadratic program using a standard SVM solver. This procedure is
guaranteed to find an $\epsilon$-approximate solution to the QP in a
polynomial number of iterations, independently of the cardinality of
the output space and of the number of examples
$n$~\cite{tsochantaridis2005large}. The $n$-slack margin rescaling
version of the CP algorithm can be found in
Algorithm~\ref{alg:cp1slack} (adapted
from~\cite{joachims2009cutting}). Please note that in our experiments
we make use of the faster, but otherwise equivalent, $1$-slack margin
rescaling variant~\cite{joachims2009cutting}. We report the
$n$-slack margin rescaling version here for ease of exposition.

\begin{algorithm}[H]
	\caption{\label{alg:cp1slack} Cutting-plane algorithm for training structural SVMs,
according to the $n$-slack formulation presented in~\cite{joachims2009cutting}.}
	\begin{small}

	\KwData{Training instances $\{(\lmtx_1,\lmty_1),\ldots,(\lmtx_n,\lmty_n)\}$, parameters $C$, $\epsilon$}
	\KwResult{Learned weights $\lmtw$}

	$\mathcal{W}_i \leftarrow \emptyset$, $\xi_i \leftarrow 0$ for all $i = 1, \ldots, n$ \;
	\Repeat{no $\mathcal{W}_i$ has changed during iteration}{
		\For{ $i = 1, \ldots, n$ }{
			$ \hat{\lmty} \leftarrow \argmax_{\hat{\lmty}} \left[ \Delta(\lmty_i,\hat{\lmty}) - \lmtw^\top \left( \lmtPsi(\lmtx_i,\lmty_i) - \lmtPsi(\lmtx_i,\hat{\lmty}) \right) \right] $ \;
			\If{ $ \Delta(\lmty_i,\hat{\lmty}) - \lmtw^\top \left( \lmtPsi(\lmtx_i,\lmty_i) - \lmtPsi(\lmtx_i,\hat{\lmty}) \right) > \lmtxi_i + \epsilon $ } {
				\begin{eqnarray*}
					\lmtw, \lmtxi	& \leftarrow	& \argmin_{\lmtw,\lmtxi \geq 0} \frac{1}{2} \lmtw^\top \lmtw + \frac{C}{n} \sum_{i=1}^n \xi_i \\
							& \text{s.t.}	& \forall {\lmty'}_1 \in \mathcal{W}_1 \;:\; \lmtw^\top \left[ \lmtPsi(\lmtx_1,\lmty_1) - \lmtPsi(\lmtx_1,{\lmty'}_1) \right] \geq \\
							&		& \qquad \Delta(\lmty_i,{\lmty'}_1) - \xi_1 \\
							&		& \vdots \\
							&		& \forall {\lmty'}_n \in \mathcal{W}_n \;:\; \lmtw^\top \left[ \lmtPsi(\lmtx_n,\lmty_n) - \lmtPsi(\lmtx_n,{\lmty'}_n) \right] \geq \\
							&		& \qquad \Delta(\lmty_i,{\lmty'}_n) - \xi_n
				\end{eqnarray*}
			}
		}
	}
	\Return{$\lmtw$}
	\end{small}
\end{algorithm}

The CP algorithm is generic, meaning that it can be adapted to any structured
prediction problem as long as it is provided with: (i) a joint feature space
representation $\lmtPsi$ of input-output pairs (and consequently a
compatibility function $f$); (ii) an oracle to perform inference, i.e. to solve
Equation~(\ref{eq:inference}); and (iii) an oracle to retrieve the most
violated constraint of the QP, i.e. to solve the {\em separation} problem:
\beq
	\argmax_{\lmty'} \lmtw^T\lmtPsi(\lmtx_i,\lmty') + \Delta(\lmtx_i,\lmty_i,\lmty')
\label{eq:separation}
\eeq
The two oracles are used as sub-routines during the optimization procedure.
For a more detailed account, and in particular for the derivation of the
separation oracle formulation, please refer to~\cite{tsochantaridis2005large}.

One key aspect of the structured output SVMs is that efficient implementations
of the two oracles are fundamental for the learning task to be tractable in
practice. The idea behind Learning Modulo Theories is that, when a hybrid
Boolean-numerical problem can be encoded in SMT, the two oracles can be
implemented using an Optimization Modulo Theory solver. This is precisely what
we propose in the present paper.
In the following sections we show how to define a feature space for hybrid
Boolean-numerical learning problems, and how to use OMT solvers to efficiently
perform inference and separation.

\subsection{Learning Modulo Theories with OMT}

Let us formalize the previous toy example in the language of LMT. In the
following we give a step-by-step description of all the building blocks of an
LMT problem: the background knowledge, the hard and soft constraints, the cost
function, and the loss function.

\paragraph{Input, Output, and Background Knowledge} Here the input
$\lmtx$ to the problem is the observed block $(x_1,y_1,dx_1,dy_1)$ while the
output $\lmty$ is the generated block $(x_2,y_2,dx_2,dy_2)$. In order to encode
the set of constraints $\{ \lmtphi_k \}$ that underlie both the learning and
the inference problems, it is convenient to first introduce a
background knowledge of predicates expressing facts about the relative
positioning of blocks. To this end we add a fresh predicate ${\tt left}(i,j)$,
that encodes the fact that ``a generic block of index $i$ touches a second
block $j$ from the left'', defined as follows:
\begin{align*}
	{\tt left}(i,j) := \;	& (x_i + dx_i = x_j) \;\land \\
				& ((y_j \le y_i \le y_j + dy_j) \;\lor\; (y_j \le y_i + dy_i \le y_j + dy_j))
\end{align*}
Similarly, we add  analogous predicates for the other directions: ${\tt
right}(i,j)$, ${\tt below}(i,j)$, ${\tt over}(i,j)$ (see Table~\ref{tab:bwbk}
for the full definitions).

\paragraph{Hard constraints}
The hard constraints represent the fact that the output $\lmty$ should be a
valid block within the bounding box (all the constraints $\lmtphi_k$ are
implicitly conjoined):
\begin{align*}
	& 0 \le x_2,y_2,dx_2,dy_2 \le 1
	& (x_2 + dx_2) \le 1 \; \land \; (y_2 + dy_2) \le 1
\end{align*}
Then we require the output block $\lmty$ to ``touch'' the input block $\lmtx$:
$$ {\tt left}(1,2) \lor {\tt right}(1,2) \lor {\tt below}(1,2) \lor {\tt over}(1,2) $$
Note that whenever this rule is satisfied, both conditions (i) and (ii) of the
toy example hold, i.e. touching blocks never overlap.

\begin{figure}[H]
	\begin{tabular}{ll}
		\hline \hline
		\multicolumn{2}{c}{ {\bf Touching blocks} } \\
		\hline
		Block $i$ touches block $j$, left	&
							{$\!
							\begin{aligned}
								{\tt left}(i,j) := \;	& x_i + dx_i = x_j \land \\
											& ((y_j \le y_i \le y_j + dy_j) \lor \\
											&  (y_j \le y_i + dy_i \le y_j + dy_j))
							\end{aligned}
							$} \\
		Block $i$ touches block $j$, below	&
							{$\!
							\begin{aligned}
								{\tt below}(i,j) := \;	& y_i + dy_i = y_j \land \\
											& ((x_j \le x_i \le x_j + dx_j) \lor \\
											&  (x_j \le x_i + dx_i \le x_j + dx_j))
							\end{aligned}
							$} \\
		Block $i$ touches block $j$, right	&
							{$\!
							\begin{aligned}
								{\tt right}(i,j) := \;	& \text{Analogous to {\tt left}$(i,j)$} \\
							\end{aligned}
							$} \\
		Block $i$ touches block $j$, over	&
							{$\!
							\begin{aligned}
								{\tt over}(i,j) := \;	& \text{Analogous to {\tt below}$(i,j)$} \\
							\end{aligned}
							$} \\
		\hline \hline
	\end{tabular}
	\caption{\label{tab:bwbk} Background knowledge used in the toy block example.}
\end{figure}

\paragraph{Cost function} Finally, we encode the cost function $ cost = w_1
dx_2 + w_2 dy_2 $, completing the description of the optimization problem.  In
the following we will see that the definition of the cost function implicitly
defines also the set of features, or equivalently the set of soft constraints,
of the LMT problem.

\paragraph{Soft constraints and Features}
Now, suppose we were given a training set of instances analogous to those
pictured in Figure~\ref{fig:bwexample}~(c), i.e. where the supervision includes
output blocks that preferentially fill as much vertical space as possible. The
learning algorithm should be able to learn this preference by inferring
appropriate weights. This kind of learning task can be cast within the
structured SVM framework, by defining an appropriate joint feature space
$\lmtPsi$ and oracles for the inference and separation problems.

Let us focus on the feature space first.
Our definition is grounded on the concept of {\em reward} assigned to an object
$(\lmtx,\lmty)$ with respect to the set of formulae $\{ \lmtphi_k \}_{k=1}^m$.
We construct the feature vector
$$ \lmtPsi(\lmtx,\lmty) := \left( \psi_1(\lmtx,\lmty), \ldots, \psi_m(\lmtx,\lmty) \right)^\top $$
by collating $m$ per-formula rewards $\psi_k(\lmtx,\lmty)$, where:
$$ \psi_k(\lmtx,\lmty) := \begin{cases} \lmtind_k(\lmtx,\lmty) &
  \text{if $\lmtphi_k$ is Boolean} \\ -c_k(\lmtx,\lmty) & \text{if
    $\lmtphi_k$ is arithmetical} \end{cases} $$
Here $\lmtind_k$ is an indicator for the satisfaction of a Boolean constraint
$\varphi_k$, while $c_k$ denotes the cost associated to real-valued
constraints, please refer to Table~\ref{tab:notation} for more details.
In other words, the feature representation of a complex object $(\lmtx,\lmty)$ is the
vector of all indicator/cost functions associated to the soft constraints.
Returning to the toy example, where the cost function is
$$ cost := w_1 \times dx_2 + w_2 \times dy_2 = \lmtw^\top \lmtPsi(\lmtx,\lmty) $$
the feature space of an instance $(\lmtx,\lmty)$ is simply
$ \lmtPsi(\lmtx,\lmty) = \left( -dx_2, -dy_2 \right) ^\top $,
which reflects the size of the output block $\lmty$. The negative sign here is
due to interpreting the features as {\em rewards} (to be maximized), while the
corresponding soft constraints can be seen as {\em costs} (to be minimized);
see Eq~\ref{eq:whatever} where this relationship is made explicit.

According to this definition both satisfied and unsatisfied rules contribute to
the total reward, and two objects $(\lmtx,\lmty)$, $(\lmtx',\lmty')$ that satisfy/violate
similar sets of constraints will be close in feature space. The compatibility
function $f(\lmtx,\lmty) := \lmtw^\top \lmtPsi(\lmtx,\lmty)$ computes the (weighted) total
reward assigned to $(\lmtx,\lmty)$ with respect to the constraints. 
Using this definition, the maximization in the inference
(Equation~\ref{eq:inference}) can be seen as attempting to find the output
$\lmty$ that maximizes the total reward with respect to the input $\lmtx$ and
the rules, or equivalently the one with minimum cost. Since $\lmtPsi$ can
be expressed in terms of  Satisfiability Modulo Linear Arithmetic, the latter
minimization problem can be readily cast as an OMT problem.
Translating back to the example, maximizing the compatibility function $f$
boils down to:
\begin{equation}
\label{eq:whatever}
\argmax \; \lmtw^\top \lmtPsi(\lmtx,\lmty) = \argmax \; ( -dx_2, -dy_2 )
\lmtw  = \argmin \; ( dx_2, dy_2 ) \lmtw
\end{equation}
which is exactly the cost minimization problem in Equation~\ref{eq:bwcost}.

\paragraph{Loss function}  The loss function determines the dissimilarity
between output structures, which in our case contain a mixture of Boolean and
rational variables.  We observe that by picking a loss function expressible as
an OMT(\larat) problem, we can readily use the same OMT solver used for
inference to also solve the CP separation oracle
(Equation~(\ref{eq:separation})). This can be achieved by selecting a loss
function such as the following Hamming loss in feature space:
\begin{eqnarray*}
	\Delta(\lmtx,\lmty,\lmty')
		& :=	& \sum_{k\;:\;\text{$\lmtphi_k$ is Boolean}} | \lmtind_k(\lmtx,\lmty) - \lmtind_k(\lmtx,\lmty') | + \\
		&	& \sum_{k\;:\;\text{$\lmtphi_k$ is arithmetical}} | c_k(\lmtx,\lmty) - c_k(\lmtx,\lmty') | \\
		& =	& \| \lmtPsi(\lmtx,\lmty) - \lmtPsi(\lmtx,\lmty') \|_1
\end{eqnarray*}
This loss function is piecewise-linear, and as such satisfies the desideratum.

Since both the inference and separation oracles required by the CP algorithm
can be encoded in OMT(\larat), we can apply an OMT solver to  efficiently solve
the learning task. In particular, our current implementation is based on a
vanilla copy of
$\text{SVM}^\text{struct}$~\footnote{http://www.cs.cornell.edu/people/tj/svm\_light/svm\_struct.html.},
which acts as a cutting-plane solver, whereas inference and separation are
implemented with the \optimathsat OMT solver.

To summarize, an LMT problem can be broken down into several components: a {\em
background knowledge}, a set of soft and hard {\em constraints}, a {\em cost
function} and a {\em loss function}. The background knowledge amounts to a set
of SMT formulae and constants useful for encoding the problem constraints,
which in turn determine the relation between inputs and outputs.  The hard
constraints define the space of candidate outputs, while the soft constraints
correspond one-to-one to features. The overall cost function is a linear
combination of the dissatisfaction/cost (or, equivalently, satisfaction/reward)
associated to the individual soft constraints, and as such is controlled
entirely by the choice of features.  Finally, the loss function determines the
dissimilarity between output structures. While in the present paper we focused
on a Hamming loss in feature space, LMT can work with any loss function that
can be encoded as an SMT formula.

\section{Experimental Evaluation} 
\label{sec:apps}

In the following we evaluate LMT on two novel that stress the ability
of LMT to deal with rather complex mixed Boolean-numerical problems.

\subsection{Stairway to Heaven}

In this section we are interested in learning how to assemble different kinds
of {\em stairways} from examples. For the purpose of this paper, a stairway is
simply a collection of $m$ blocks (rectangles) located within a
two-dimensional, unit-sized bounding box
$[0,1]\times[0,1]$.
Clearly not all possible arrangements of blocks form a stairway; a stairway
must satisfy the following conditions: (i) the first block touches either the
top or the bottom corner of the left edge of the bounding box; (ii) the last
block touches the opposite corner at the right edge of the bounding box; (iii)
there are no gaps between consecutive blocks; (iv) consecutive blocks must
actually form a step and, (v) no two blocks overlap.
Note that the property of ``being a stairway'' is a {\em collective} property
of the $m$ blocks.

More formally, each block $i=1,\ldots,m$ consists of four rational variables:
the origin $(x_i,y_i)$, which indicates the bottom-left corner of the block, a
width $dx_i$ and a height $dy_i$; the top-right corner of the block is
$(x_i+dx_i,y_i+dy_i)$. A stairway is simply an assignment to all $4\times m$
variables that satisfies the above conditions.

\begin{figure}[H]
	\centering
	\includegraphics{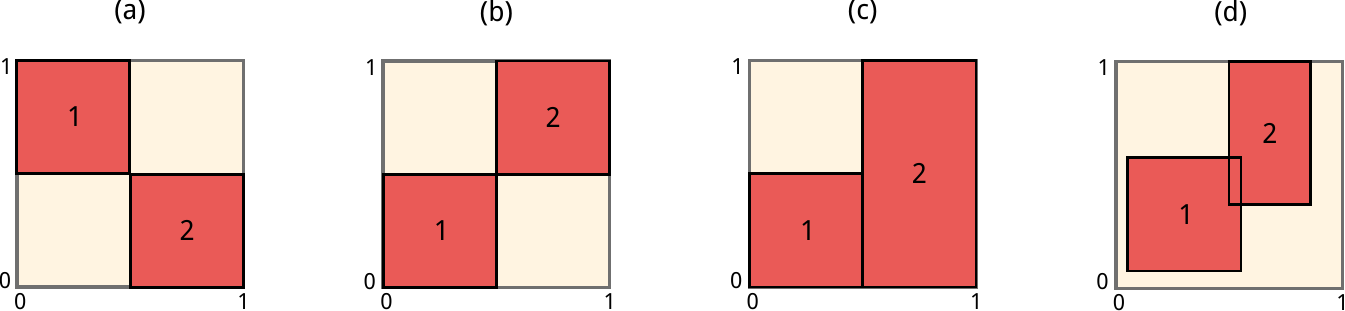}
	\caption{\label{fig:stairwayexample} (a) A {\em left ladder} $2$-stairway. (b) A {\em right ladder}
$2$-stairway. (c) A {\em right pillars} $2$-stairway.  (d) A block assignment
that violates conditions (i), (ii) and (iv), and as such does not form a
stairway.}
\end{figure}

Our definition does not impose any constraint on the {\em orientation} of
stairways: it is perfectly legitimate to have {\em left} stairways that start
at the top-left corner of the bounding box and reach the bottom-right corner,
and {\em right} stairways that connect the bottom-left corner to the top-right
one.
For instance, a left $2$-stairway can be defined with the following block
assignment (see Figure~\ref{fig:stairwayexample}~(a)):
$$ (x_1,y_1,dx_1,dy_1) = \left(0,\frac{1}{2},\frac{1}{2},\frac{1}{2}\right) \qquad (x_2,y_2,dx_2,dy_2) = \left(\frac{1}{2},0,\frac{1}{2},\frac{1}{2}\right) $$
Similarly, a right $2$-stairway is obtained with the assignment
(Figure~\ref{fig:stairwayexample}~(b)):
$$ (x_1,y_1,dx_1,dy_1) = \left(0,0,\frac{1}{2},\frac{1}{2}\right) \qquad (x_2,y_2,dx_2,dy_2) = \left(\frac{1}{2},\frac{1}{2},\frac{1}{2},\frac{1}{2}\right) $$

We also note that the above conditions do not impose any explicit restriction
on the width and height of individual blocks (as long as consecutive ones are
in contact and there is no overlap). Consequently we allow for both {\em
ladder} stairways, where the total amount of vertical and horizontal surface of
the individual blocks is minimal, as in
Figure~\ref{fig:stairwayexample}~(a) and (b); and for {\em pillar} stairways,
where either the vertical or horizontal block lengths are maximized, as in
Figure~\ref{fig:stairwayexample}~(c).
There are of course an uncountable number of intermediate stairways that do not
belong to any of the above categories.

Inference amounts to generating a set of variable assignments to all blocks, so
that none of conditions (i)-(v) is violated, and the cost of the soft rules
is minimized. This can be easily encoded as an OMT(\larat) problem.
As a first step, we define a background knowledge of useful predicates. We use
four predicates to encode the fact that a block $i$ may touch one of the four
corners of the bounding box, namely ${\tt bottom\_left}(i)$, ${\tt
bottom\_right}(i)$, ${\tt top\_left}(i)$, and ${\tt top\_right}(i)$, which can
be written as, e.g.:
$$ {\tt bottom\_right}(i) := (x_i + dx_i) = 1 \land y_i = 0 $$
We also define predicates to describe the relative positions of two blocks $i$
and $j$, such as ${\tt left}(i,j)$:

\begin{align*}
	{\tt left}(i,j) := \;	& (x_i + dx_i) = x_j \land \\
				& ((y_j \le y_i \le y_j + dy_j) \lor \\
				&  (y_j \le y_i + dy_i \le y_j + dy_j))
\end{align*}
encodes the fact that block $i$ is touching block $j$ from the
left. Similarly, we also define ${\tt below}(i,j)$ and
${\tt over}(i,j)$.  Finally, and most importantly, we combine the
above predicates to define the concept of {\em step}, i.e. two blocks
$i$ and $i+1$ that are both touching and positioned as to form a
stair:
\begin{align*}
	{\tt left\_step}(i,j) := \;	& ({\tt left}(i,j) \land (y_i + dy_i) > (y_j + dy_j)) \; \lor \\
					& ({\tt over}(i,j) \land (x_i + dx_i) < (x_j + dx_j))
\end{align*}
We define {\tt right\_step}(i,j) in the same manner. For a complete description
of the background knowledge, see Table~\ref{tab:sthbk}.

\begin{table}[H]
	\begin{small}
	\begin{tabular}{ll}
		\hline \hline
		\multicolumn{2}{c}{ {\bf Corners} } \\
		\hline
		Block $i$ at bottom-left corner		& $ {\tt bottom\_left}(i) := x_i = 0 \land y_i = 0 $ \\
		Block $i$ at bottom-right corner	& $ {\tt bottom\_right}(i) := (x_i + dx_i) = 1 \land y_i = 0 $ \\
		Block $i$ at top-left corner		& $ {\tt top\_left}(i) := x_i = 0 \land (y_i + dy_i) = 1 $ \\
		Block $i$ at top-right corner		& $ {\tt top\_right}(i) := (x_i + dx_i) = 1 \land (y_i + dy_i) = 1 $ \\
		\hline \hline
		\multicolumn{2}{c}{ {\bf Relative block positions} } \\
		\hline
		Block $i$ touches block $j$, left		&
							{$\!
							\begin{aligned}
								{\tt left}(i,j) := \;	& (x_i + dx_i) = x_j \land \\
											& ((y_j \le y_i \le y_j + dy_j) \lor \\
											&  (y_j \le y_i + dy_i \le y_j + dy_j)) \\
							\end{aligned}
							$} \\
		Block $i$ touches block $j$, below           & 
							{$\!
							\begin{aligned}
								{\tt below}(i,j) := \;	& (y_i + dy_i) = y_j \land \\
											& ((x_j \le x_i \le x_j + dx_j) \lor \\
											&  (x_j \le x_i + dx_i \le x_j + dx_j))
							\end{aligned}
							$} \\
		Block $i$ touches block $j$, over           &
							{$\!
							\begin{aligned}
								{\tt over}(i,j) := \;	& (y_j + dy_j) = y_i \land \\
											& ((x_j \le x_i \le x_j + dx_j) \lor \\
											&  (x_j \le x_i + dx_i \le x_j + dx_j))
							\end{aligned}
							$} \\
		\hline \hline
		\multicolumn{2}{c}{ {\bf Steps} } \\
		\hline
		Left step				&
							{$\!
							\begin{aligned}
								& {\tt left\_step}(i,j) := \; \\
                                                                & \qquad ({\tt left}(i,j) \land (y_i + dy_i) > (y_j + dy_j)) \; \lor \\
								& \qquad ({\tt over}(i,j) \land (x_i + dx_i) < (x_j + dx_j))
							\end{aligned}
							$} \\
		Right step				&
							{$\!
							\begin{aligned}
								& {\tt right\_step}(i,j) := \;	\\
                                                                & \qquad ({\tt left}(i,j)  \land (y_i + dy_i) < (y_j + dy_j)) \; \lor \\
                                                                & \qquad ({\tt below}(i,j) \land (x_i + dx_i) < (x_j + dx_j))
							\end{aligned}
							$} \\
		\hline \hline
	\end{tabular}
	\end{small}
	\caption{\label{tab:sthbk} Background knowledge used in the stairways experiment.}
\end{table}

The background knowledge allows to encode the property of being a {\em left}
stairway as:
$$ {\tt top\_left}(1) \land \bigwedge_{i\in[1,m-1]} \; {\tt left\_step}(i,i+1) \land {\tt bottom\_right}(m) $$
Analogously, any {\em right} stairway satisfies the following condition:
$$ {\tt bottom\_left}(1) \land \bigwedge_{i\in[1,m-1]} \; {\tt right\_step}(i,i+1) \land {\tt top\_right}(m) $$
However, our inference procedure does not have access to this knowledge. We
rather encode an appropriate set of {\em soft rules} (costs) which, along with
the associated weights, should bias the optimization towards block assignments that
form a stairway of the correct type.

We include a few hard rules to constrain the space of admissible block
assignments. We require that all blocks fall within the bounding box:
$$ \forall i \; 0 \le x_i,dx_i,y_i,dy_i \le 1 $$
$$ \forall i \; 0 \le (x_i+dx_i) \le 1 \; \land \; 0 \le (y_i+dy_i) \le 1 $$
We also require that blocks do not overlap:
\begin{eqnarray*}
	\forall i \neq j & & (x_i + dx_i \le x_j) \lor (x_j + dx_j \le x_i) \lor \\
                         & & (y_i + dy_i \le y_j) \lor (y_j + dy_j \le y_i)
\end{eqnarray*}
Finally, we require (without loss of generality) blocks to be ordered from
left to right, $ \forall i \; x_i \leq x_{i+1} $.

Note that only condition (v) is modelled as a hard constraint. The others
are implicitly part of the problem {\em cost}. Our cost model is based on the
observation that it is possible to discriminate between the different stairway
types using only four factors: minimum and maximum step size, and amount of
horizontal and vertical {\em material}. These four factors are useful features
in discriminating between the different stairway types without having to resort
to quadratic terms, e.g. the areas of the individual blocks.
For instance, in the cost we account for both the maximum step height of all
left steps (a good stairway should not have too high steps):
$$ maxshl = m \times \max_{i\in[1,m-1]} \begin{cases} (y_i+dy_i) - (y_{i+1}+dy_{i+1}) & \text{if $i,i+1$ form a left step} \\ 1 & \text{otherwise} \end{cases} $$
and the minimum step width of all right steps (good stairways should have
sufficiently large steps):
$$ minswr = m \times \min_{i\in[1,m-1]} \begin{cases} (x_{i+1}+dx_{i+1}) - (x_i+dx_i) & \text{if $i,i+1$ form a right step} \\ 0 & \text{otherwise} \end{cases} $$
The value of these costs depends on whether a pair of blocks actually
forms a left step, a right step, or no step at all. Note that these
costs are multiplied by the number of blocks $m$. This allows to
renormalize costs according to the number of steps; e.g. the step
height of a stairway with $m$ uniform steps is half that of a stairway with
$m/2$ steps. 
Finally, we write the average amount of vertical material as $vmat =
\frac{1}{m} \sum_i dy_i$.  All the other costs can be written
similarly; see Table~\ref{tab:sthrules} for the complete list.  As we
will see, the normalization of individual costs allows to learn
weights which generalize to stairways with a larger number of blocks with
respect to those seen during training. 

Putting all the pieces
together, the complete cost is:
\begin{eqnarray*}
	cost	& :=	& (  maxshl, minshl, maxshr, minshr, \\
		&	& \; maxswl, minswl, maxswr, minswr, \\
		&	& \; vmat,   hmat ) \; \lmtw
\end{eqnarray*}

Minimizing the weighted cost implicitly requires the inference engine to decide
whether it is preferable to generate a {\em left} or a {\em right} stairway,
thanks to the $minshl,\ldots,minswr$ components, and whether the stairway
should be a {\em ladder} or {\em pillar}, due to $vmat$ and $hmat$.
The actual weights are learned, allowing the learnt model to reproduce
whichever stairway type is present in the training data.

\begin{table}[H]
	\centering
	\begin{tabular}{ll}
		\hline
		\hline
		\multicolumn{2}{c}{ {\bf (a) Hard constraints} } \\
		\hline
		Bounding box			&
						{$\!
						\begin{aligned}
							\forall i \; & 0 \leq x_i + dx_i \leq 1 \; \land \\
						                     & 0 \leq y_i + dy_i \leq 1
						\end{aligned}
						$} \\
		[1.2em]
		No overlap			&
						{$\!
						\begin{aligned}
							\forall i \neq j \; & (x_i + dx_i \le x_j) \lor (x_j + dx_j \le x_i) \lor \\
						                            & (y_i + dy_i \le y_j) \lor (y_j + dy_j \le y_i)
						\end{aligned}
						$} \\
		[1.2em]
		Blocks left to right		& $ \forall i \; x_i \leq x_{i+1} $ \\
		[0.6em]
		\hline
		\hline
		\multicolumn{2}{c}{ {\bf (b) Soft constraints (features)} } \\
		\hline
		Max step height left		& $ maxshl = m \times \max_i (y_i+dy_i) - (y_{i+1}+dy_{i+1}) $ \\
		Min step height left		& $ minshl = m \times \min_i (y_i+dy_i) - (y_{i+1}+dy_{i+1}) $ \\
		Max step height right		& $ maxshr = m \times \max_i (y_{i+1}+dy_{i+1}) - (y_i+dy_i) $ \\
		Min step height right		& $ minshr = m \times \min_i (y_{i+1}+dy_{i+1}) - (y_i+dy_i) $ \\
		[1.2em]
		Max step width left		& $ maxswl = m \times \max_i (x_{i+1}+dx_{i+1}) - (x_i+dx_i) $ \\
		Min step width left		& $ minswl = m \times \min_i (x_{i+1}+dx_{i+1}) - (x_i+dx_i) $ \\
		Max step width right		& $ maxswr = m \times \max_i (x_i+dx_i) - (x_{i+1}+dx_{i+1}) $ \\
		Min step width right		& $ minswr = m \times \min_i (x_i+dx_i) - (x_{i+1}+dx_{i+1}) $ \\
		[1.2em]
		Vertical material		& $ vmat = \frac{1}{m} \sum_i dy_i $ \\
		[0.2em]
		Horizontal material		& $ hmat = \frac{1}{m} \sum_i dx_i $ \\
		[0.6em]
		\hline
		\hline
		\multicolumn{2}{c}{ {\bf (c) Cost} } \\
		\hline
		\multicolumn{2}{c}{
			{$\!
			\begin{aligned}
				cost :=  (	& maxshl, minshl, maxshr, minshr, \\
							& maxswl, minswl, maxswr, minswr, \\
							& vmat,   hmat
                                                        ) \; \lmtw
			\end{aligned}
			$}
		} \\
		\hline
		\hline
	\end{tabular}
	\caption{\label{tab:sthrules} List of all rules used in the stairway problem. Top, hard
rules affect both the inference and separation procedures. Middle, soft rules
(costs), whose weight is learned from data. Bottom, the total cost of a
stairway instance is the weighted sum of all individual soft constraints.}
\end{table}

To test the stairway scenario, we focused on learning one model for each of
six kinds of stairway: {\em left ladder}, {\em right ladder}, {\em left pillar} and
{\em right pillar} with a preference for horizontal blocks, and {\em left
pillar} and {\em right pillar} with vertical blocks. In this setting, the input
$\lmtx$ is empty, and the model should generate all $m$ blocks as
output $\lmty$ during test.

\begin{figure}[H]
	\includegraphics{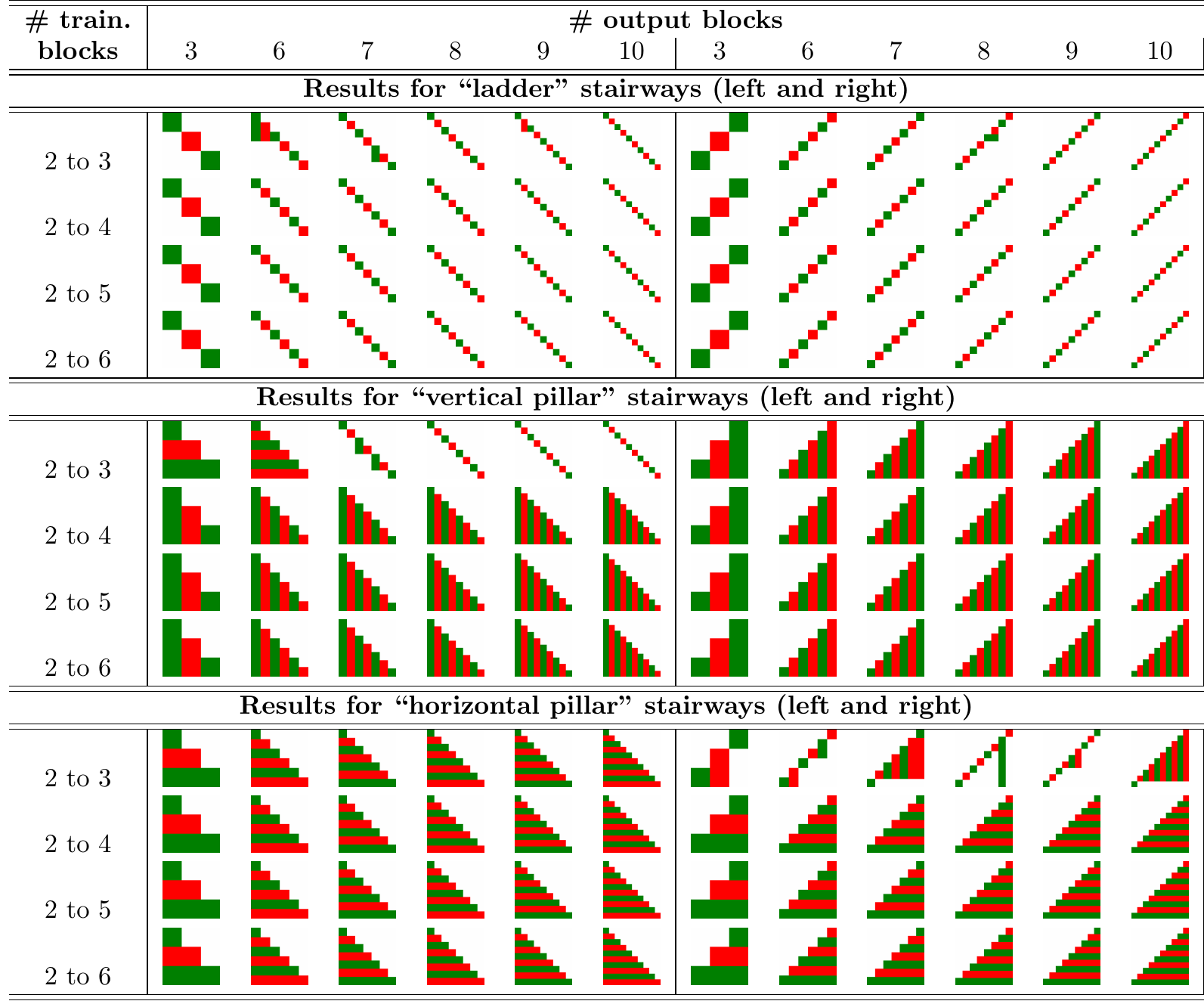}
		\caption{\label{fig:sthresults} Results for the stairway construction problem. From top to
bottom: results for the {\em ladder}, {\em horizontal pillar}, and {\em
vertical pillar} cases. Images in row labeled with $N$ picture the stairways
generated by a model learnt on a training set of perfect stairways made of
$2,3,\ldots,N$ steps, with a varying number of generated blocks.}
\end{figure}

We generated ``perfect'' stairways of $2$ to $6$ blocks for each stairway type
to be used as training instances.
We then learned a model using all training instances up to a fixed number of
blocks: a model using examples with up to $3$ blocks, another with examples of
up to $4$, {\em etc.}, for a total of $4$ models per stairway type.
Then we analyzed the generalization ability of the learnt models by generating
stairways with a {\em larger} number of blocks (up to $10$) than those in the
training set.
The results can be found in Figure~\ref{fig:sthresults}.

The experiment shows that LMT is able to solve the stairway
construction problem, and can learn appropriate models for all
stairway types, as expected.
As can be seen in~Figure~\ref{fig:sthresults}, the generated stairways can
present some imperfections when the training set is too small (e.g., only two
training examples; first row of each table), especially in the $10$ output
blocks case.
However, the situation quickly improves when the training set
increases: models learned with four training examples are always able
to produce perfect $10$-block stairways of the same kind. Note again
that the learner has no explicit notion of what a stairway is, but just
the values of step width, height and material for some training
examples of stairways. 

All in all, albeit simple, this experiment showcases the ability of
LMT to handle learning in hybrid Boolean-numerical domains, whereas
other formalisms are not particularly suited for the task.
As previously mentioned, the Church~\cite{church08} language allows to
encode arbitrary constraints over both numeric and Boolean
variables. The stairway problem can indeed be encoded in Chuch in a
rather straightforward way. However, the sampling strategies used for
inference are not conceived for performing optimization with hard
continuous constraints. Even the simple task of generating two blocks,
conditioned on the fact that they form a step, is prohibitively
expensive.~\footnote{We interrupted the inference process after 24 hours
of computation.}
%

\subsection{Learning to Draw Characters}

In this section we are concerned with automatic character drawing, a
novel structured-output learning problem that consists in learning to
translate any input noisy hand-drawn character into its symbolic
representation.
More specifically, given an unlabeled black-and-white image of a handwritten
letter or digit, the goal is to construct an equivalent {\em vectorial}
representation of the same character.

In this paper, we assume the character to be representable by a
polyline made of a given number $m$ of {\em directed} segments,
i.e. segments identified by a starting point $(x^b,y^b)$ and an ending
point $(x^e,y^e)$.
The input image $\lmtx$ is seen as the set $P$ of coordinates of the pixels
belonging to the character, while the output $\lmty$ is a set of $m$ directed
segments
$ \{ (x^b_i,y^b_i,x^e_i,y^e_i) \}_{i=1}^m $. 
Just like in the previous section, we assume all coordinates to fall within
the unit bounding box.

Intuitively, any good output $\lmty$ should satisfy the following requirements:
(i) it should be as similar as possible to the noisy input character; and (ii)
it should actually ``look like'' the corresponding vectorial character.
Here we interpret the first condition as  implying that the generated segments
should cover as many pixels of the input image as possible (although
alternative interpretations are possible).
Under this interpretation, we can informally write the inference problem as
follows:
$$ \argmax_{\lmty} \; ( coverage(\lmtx,\lmty), orientation(\lmty) )
\;  \lmtw $$
where the $orientation$ term encodes information on the orientation of
the segments which should be useful for computing the ``looking like''
condition. In the following, we will detail how to formulate and
compute these two quantities.

\begin{figure}[H]
	\centering
	\includegraphics{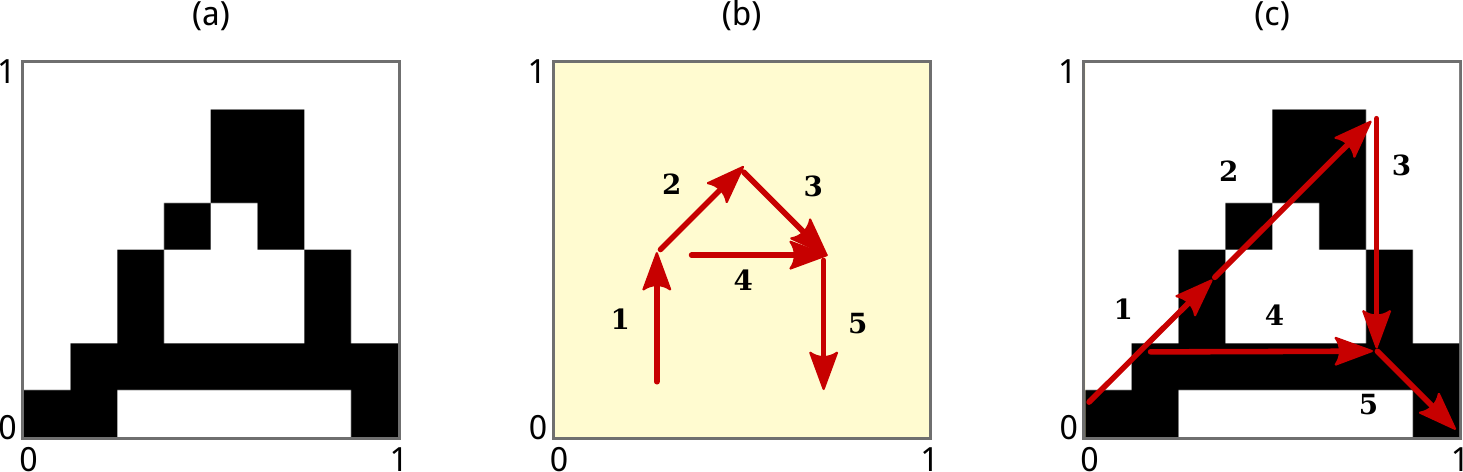}
		\caption{\label{fig:charexample} Left, example $8\times8$ bitmap image of an ``A''. Middle,
a set of 5 segments satisfying the ``looking like an A'' rules in the text.
Right, 5 segments satisfying both the rules for character ``A'' and fitting
the underlying image.}
\end{figure}

Since the output is supposed to be a polyline, we constrain consecutive
segments to be connected, i.e. to share an endpoint:
$$ \forall i \; {\tt connected}(i,i+1) $$
We also want the segments to be no larger than the image, nor smaller than
a pixel:
$\forall i \; min\_length \le length(i) \le 1$.
Additionally, we constrain (without loss of generality) each segment
to be ordered from left to right, i.e. $x^b_i \le x^e_i$.
Finally, we restrict the segments
to be either horizontal, vertical or $45^\circ$ diagonal, that is:
$$ \forall i \; {\tt horizontal}(i) \lor {\tt vertical}(i) \lor {\tt diagonal}(i) $$
This restriction allows us express all numerical constraints in linear
terms.
All the predicates used above are defined as in Table~\ref{tab:charbk}.

Under these assumptions, we can encode the $coverage$ reward as:
$$ 
coverage(\lmtx,\lmty) := \frac{1}{|P|} \sum_{p\in P} \lmtind({\tt covered}(p)) $$
where ${\tt covered}(p)$ is true if pixel $p$ is covered by at
least one of the $m$ segments:
$$
{\tt covered}(p) := \bigvee_{i \in [1,m]} {\tt covered}(p,i)
$$
The fact that a segment $i=(x^b_i,y^b_i,x^e_i,y^e_i)$ covers pixel
$p=(x,y)$ implicitly depends on the orientation of the segment, and is
computed using constructs like:
$$
\textrm{If} \quad {\tt horizontal}(i) \quad \textrm{then} \quad {\tt covered}(p,i) := x^b_i \le x \le x^e_i \land y = y^b_i 
$$
The coverage formulae for the other segment types can be found in
Table~\ref{tab:charbk}.
	
As for the $orientation$ term, it should contain features related to
the vectorial representation of characters. These include both the
direction of the individual segments and the connections between pairs
of segments. As an example, consider this possible description of
``looking like an A'':
\begin{eqnarray*}
	 {\tt increasing}(1) & \;\land\; & {\tt h2t}(1,2) \;\land\; \\
	 {\tt increasing}(2) & \;\land\; & {\tt h2t}(2,3) \;\land\; \\
	 {\tt decreasing}(3) & \;\land\; & {\tt h2h}(3,4) \;\land\; \\
	 {\tt horizontal}(4) & \;\land\; & {\tt h2t}(4,5) \;\land\; \\
	 {\tt decreasing}(5) & &
\end{eqnarray*}
Here ${\tt increasing}(i)$ and ${\tt decreasing}(i)$ indicate the direction of
segment $i$, and can be written as:
\begin{eqnarray*}
	{\tt increasing}(i)	& :=	& y_i^e > y_i^b \\
	{\tt decreasing}(i)	& :=	& y_i^e < y_i^b
\end{eqnarray*}
Then we have connections between pairs of segments. We encode
connection types following the convention used for Bayesian Networks,
where the head of a directed segment is the edge containing the arrow
(represented by the ending point $(x^e,y^e)$) and the tail is the opposite edge
(the starting point $(x^b,y^b)$). For instance, ${\tt h2t}(i,j)$ indicates
that $i$ is head-to-tail with respect to $j$, ${\tt h2h}(i,j)$ that they are
head-to-head:
\begin{eqnarray*}
	{\tt h2t}(i,j)	& :=	& (x_i^e = x_j^b) \land (y_i^e = y_j^b) \\
	{\tt h2h}(i,j)	& :=	& (x_i^e = x_j^e) \land (y_i^e = y_j^e)
\end{eqnarray*}
For a pictorial representation of the ``looking like an A''
constraint, see Figure~\ref{fig:charexample}~(b). We include a number
of other, similar predicates in the background knowledge; for a full
list, see Table~\ref{tab:charbk}.

For example, suppose we have a $8\times8$ image of an upper-case ``A'', as in
Figure~\ref{fig:charexample}~(a). A character drawing algorithm should decide
how to overlay $5$ segments on top of the input image according to the previous
two criteria. A good output would look like the one pictured in
Figure~\ref{fig:charexample}~(c).

However, the formula for the ``looking like an A'' constraint is not
available at test time and should be learned from the data. In order
to do so, the $orientation$ term includes possible directions ({\tt
  increasing}, {\tt decreasing}, {\tt right}) for all $m$ segments and
all possible connection types between {\em all} segments ({\tt
  h2t}, {\tt h2h}, {\tt t2t}, {\tt t2h}).
Note that we do not include detailed segment orientation (i.e., {\tt
horizontal}, {\tt vertical}, {\tt diagonal}) in the feature space to
accommodate for alternative vectorial representations of the same letter. For
instance, the first segment in an ``A'', which due to the left-to-right rule
necessarily fits the lower left portion of the character, is bound to be {\tt
increasing}, but may be equally likely {\tt vertical} or {\tt
diagonal} (see e.g. Figures~\ref{fig:charexample}~(b) and~(c)). 

Summing up, the $orientation$ term can be written as:
\begin{eqnarray*}
&(&increasing(1), decreasing(1), right(1), \\
& & \cdots \\
& & increasing(m), decreasing(m), right(m), \\
& & h2t(1,2), t2h(1,2), h2h(1,2), t2t(1,2), \\
& & \cdots \\
& & h2t(1,m), t2h(1,m), h2h(1,m), t2t(1,m), \\
& & \cdots \\
& & h2t(m-1,m), t2h(m-1,m), h2h(m-1,m), 
          t2t(m-1,m)\;\;)
\end{eqnarray*}
where each feature is the indicator function of the corresponding
Boolean variable, e.g. $increasing(1) := \lmtind({\tt increasing}(1))$
 (see Table~\ref{tab:charrules}).

\begin{table}[H]
	\begin{tabular}{ll}
		\hline \hline
		\multicolumn{2}{c}{ {\bf Segment types} } \\
		\hline
		Segment $i$ is horizontal		& $ {\tt horizontal}(i) := (x_i^b \ne x_i^e) \land (y = y_i^b) $ \\
		Segment $i$ is vertical			& $ {\tt vertical}(i) := (x_i^b = x_i^e) \land (y_i^e \ne y_i^b) $ \\
		Segment $i$ is diagonal			& $ {\tt diagonal}(i) := |x_i^e-x_i^b|  = |y_i^e - y_i^b| $ \\
		Segment $i$ is increasing		& $ {\tt increasing}(i) := y_i^e > y_i^b $ \\
		Segment $i$ is decreasing		& $ {\tt decreasing}(i) := y_i^e < y_i^b $ \\
		Segment $i$ is left-to-right		& $ {\tt right}(i) := x_i^e > x_i^b $ \\
		Segment $i$ is incr. vert.		& $ {\tt
                  incr\_vert}(i) := {\tt increasing}(i) \land {\tt
                  vertical}(i) $\\
		Segment $i$ is decr. vert.		& $ {\tt
                  decr\_vert}(i) :=  {\tt decreasing}(i) \land {\tt
                  vertical}(i) $\\
		Segment $i$ is incr. diag.		& $ {\tt
                  incr\_diag}(i) := {\tt increasing}(i) \land {\tt
                  diagonal}(i) $\\
		Segment $i$ is decr. diag.		& $ {\tt
                  decr\_diag}(i) :=  {\tt decreasing}(i) \land {\tt
                  diagonal}(i) $\\
		\hline \hline
		\multicolumn{2}{c}{ {\bf Segment length} } \\
		\hline
		Lenght of horiz. segment $i$		& $ {\tt horizontal}(i) \to length(i) = |x_i^e-x_i^b| $ \\
		Length of vert. segment $i$		& $ {\tt vertical}(i) \to length(i) = |y_i^e-y_i^b| $ \\
		Lenght of diag. segment $i$		& $ {\tt diagonal}(i) \to length(i) = \sqrt{2} \, |y_i^e-y_i^b| $ \\
		\hline \hline
		\multicolumn{2}{c}{ {\bf Connections between segments} } \\
		\hline
		Segments $i$,$j$ are head-to-tail	& $ {\tt h2t}(i,j) := (x_i^e = x_j^b) \land (y_i^e = y_j^b) $ \\
		Segments $i$,$j$ are head-to-head	& $ {\tt h2h}(i,j) := (x_i^e = x_j^e) \land (y_i^e = y_j^e) $ \\
		Segments $i$,$j$ are tail-to-tail	& $ {\tt
                  t2t}(i,j) := (x_i^b = x_j^b) \land (y_i^b = y_j^b) $
                \\
		Segments $i$,$j$ are tail-to-head	& $ {\tt t2h}(i,j) := (x_i^b = x_j^e) \land (y_i^b = y_j^e) $ \\
		Segments $i$,$j$ are connected		&
							{$\!
							\begin{aligned}
								{\tt connected}(i,j) := \;	& {\tt h2h}(i,j) \lor {\tt h2t}(i,j) \lor \\
												& {\tt t2h}(i,j) \lor {\tt t2t}(i,j)
							\end{aligned}
							$} \\
		\hline \hline
		\multicolumn{2}{c}{ {\bf Whether segment $i = (x^b,y^b,x^e,y^e)$ covers pixel $p = (x,y)$} } \\
		\hline
                Coverage of pixel $p$		& $ {\tt covered}(p) := \bigvee_i {\tt covered}(p,i) $ \\
		Coverage of pixel $p$ by seg. $i$	& \\

		\multicolumn{2}{l}{\qquad {\tt incr\_vert}$(i)$ $\to$  ${\tt covered}(p,i)
                := y^b_i \le y \le y^e_i \land x = x^b_i$} \\ 
		\multicolumn{2}{l}{\qquad{\tt decr\_vert}$(i)$ $\to$  ${\tt covered}(p,i)
                := y^e_i \le y \le y^b_i \land x = x^b_i $}\\
		\multicolumn{2}{l}{\qquad{\tt horizontal}$(i)$ $\to$  ${\tt covered}(p,i)
                := x^b_i \le x \le x^e_i \land y = y^b_i $}\\                
		\multicolumn{2}{l}{\qquad{\tt incr\_diag}$(i)$ $\to$  ${\tt covered}(p,i)
                := y^b_i \le y \le y^e_i \land x^b_i \le x \le x^e_i
                \land x^b_i - y^b_i = x-y$} \\
		\multicolumn{2}{l}{\qquad {\tt decr\_diag}$(i)$ $\to$  ${\tt covered}(p,i)
                := y^e_i \le y \le y^b_i \land x^b_i \le x \le x^e_i
                \land x^b_i + y^b_i = x+y$}\\
	
		\hline \hline
	\end{tabular}
	\caption{\label{tab:charbk} Background knowledge used in the character writing experiment.}
\end{table}

\begin{table}[H]
	\centering
	\begin{tabular}{ll}
		\hline
		\hline
		\multicolumn{2}{c}{ {\bf (a) Hard constraints} } \\
		\hline
		Left-to-right ordering			& $ x_i^b \le x_i^e $ \\
		Allowed segment types			& $ {\tt vertical}(i) \lor {\tt horizontal}(i) \lor {\tt diagonal}(i) $ \\
		Consecutive segments are connected	& $ {\tt connected}(i,i+1) $ \\
		Minimum segment size			& $ min\_length \le length(i) \le 1 $ \\
		\hline
		\hline
		\multicolumn{2}{c}{ {\bf (b) Soft constraints (features)} } \\
		\hline
		Non-zero pixel coverage 		& $ coverage
                := \frac{1}{|P|} \sum_{p \in P} \lmtind({\tt covered}(p)) $ \\
		Indicator of increasing segment $i$	& $ increasing(i) := \lmtind({\tt increasing}(i)) $ \\
		Indicator of decreasing segment $i$	& $ decreasing(i) := \lmtind({\tt decreasing}(i)) $ \\
		Indicator of right segment $i$		& $ right(i) := \lmtind({\tt right}(i)) $ \\
		Indicator of head-to-tail $i$, $j$	& $ h2t(i,j) := \lmtind({\tt h2t}(i,j)) $ \\
		Indicator of tail-to-head $i$, $j$	& $ t2h(i,j) := \lmtind({\tt t2h}(i,j)) $ \\
		Indicator of head-to-head $i$, $j$	& $ h2h(i,j) := \lmtind({\tt h2h}(i,j)) $ \\
		Indicator of tail-to-tail $i$, $j$	& $ t2t(i,j) := \lmtind({\tt t2t}(i,j)) $ \\
		\hline
		\multicolumn{2}{c}{ {\bf (c) Cost} } \\
		\hline
		\multicolumn{2}{c}{
			{$\!
			\begin{aligned}
				cost := \lmtw^\top (	& \underbrace{increasing(i), decreasing(i), right(i)}_{\text{for all segments $i$}}, \\
							& \underbrace{h2t(i,i+1), t2h(i,i+1), h2h(i,i+1), t2t(i,i+1)}_{\text{for all segments $i$}}, \\
							& coverage ) \\
			\end{aligned}
			$}
		} \\
		\hline
		\hline
	\end{tabular}
	\caption{\label{tab:charrules} List of all rules used in the character writing problem. Top,
hard rules. Middle, soft rules (costs). Bottom, total cost of a segment
assignment.}
\end{table}

\begin{figure}[H]
	\includegraphics{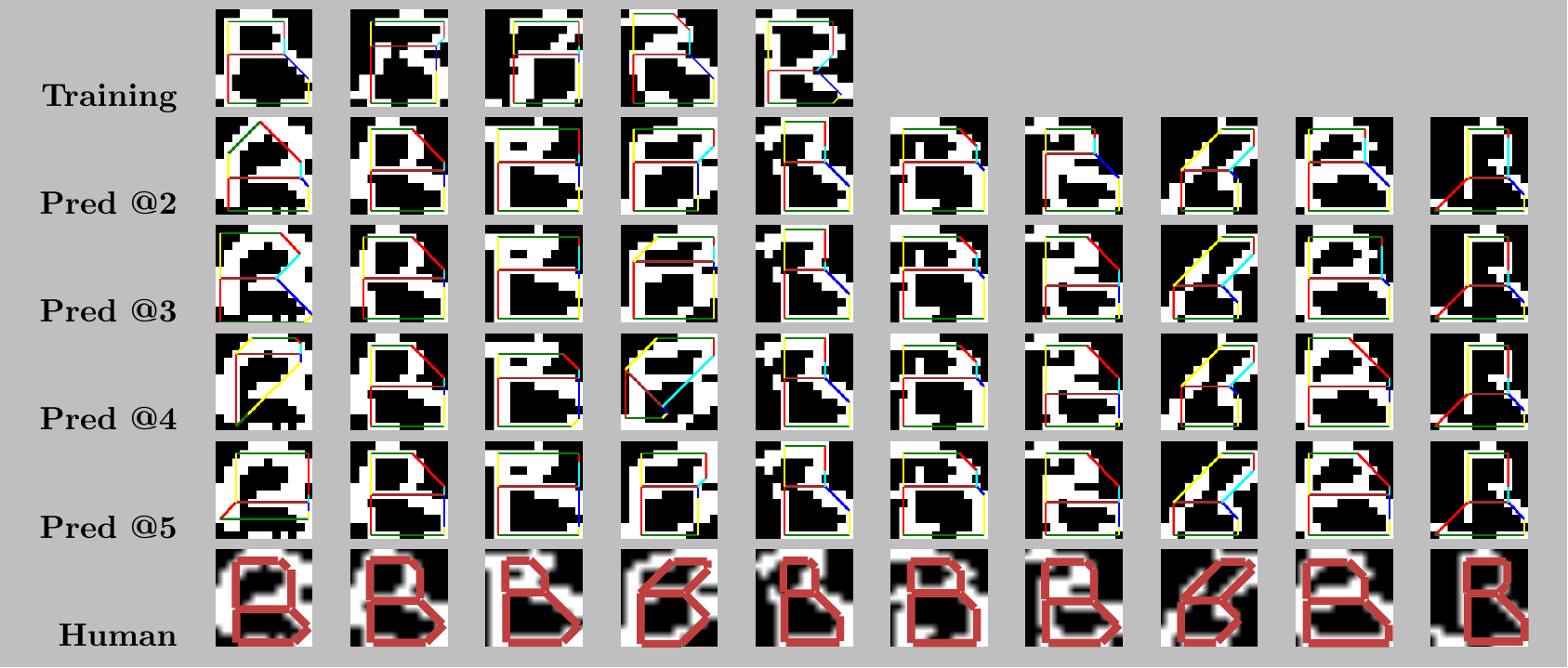}
	\caption{\label{fig:charB} Results for the ``B'' $12\times12$ character
drawing task.}
\end{figure}

We evaluated LMT on the character drawing problem by carrying out an
extensive experiment using a set of noisy B\&W $16\times20$ character
images\footnote{Dataset taken from
  http://cs.nyu.edu/$\sim$roweis/data.html}. The dataset includes $39$
instances of handwritten images of each alphanumerical character. We
downscaled the images to $12\times12$ for speeding up the experiments.
Learning to draw characters is a very challenging constructive problem, made
even more difficult by the low quality of the noisy images in the dataset
(see, e.g. Figure~\ref{fig:charB}.)
In this experiment we learn a model for each of the first five letters
of the alphabet (A to E), and assess the ability of LMT to generalize
over unseen handwritten images of the same character.

\begin{figure}[H]
	\includegraphics{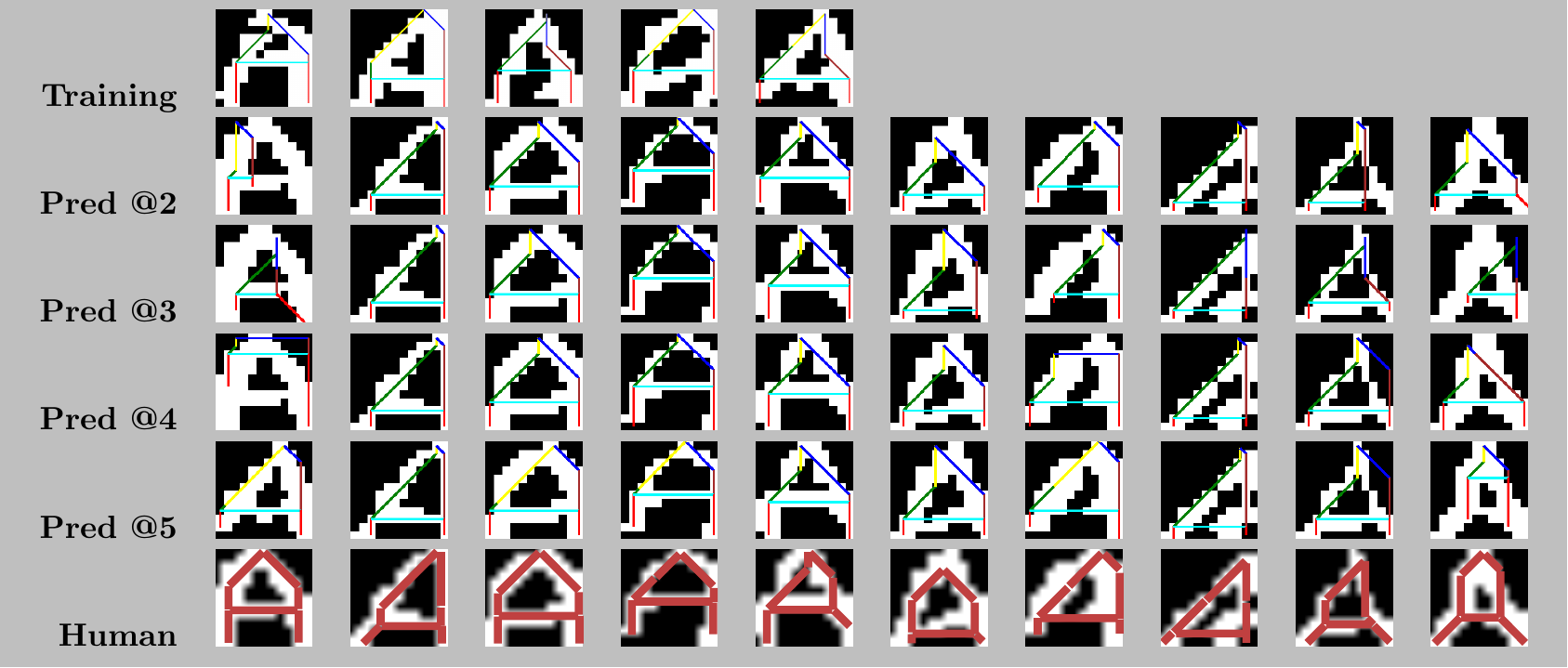}
	\caption{\label{fig:charA} Results for the ``A'' $12\times12$ character
drawing task. The training instances are lined up in the first row. The second
to fifth row are the segmentations generated by models learned with the first
two training instances, the first three instances, {\em etc.}, respectively.
The last row are the human-made segmentations used in the comparison. The
generated vectorial representations are shown overlayed over the corresponding
bitmap image.  Segments are colored for readability.}
\end{figure}

\begin{figure}[H]
	\includegraphics{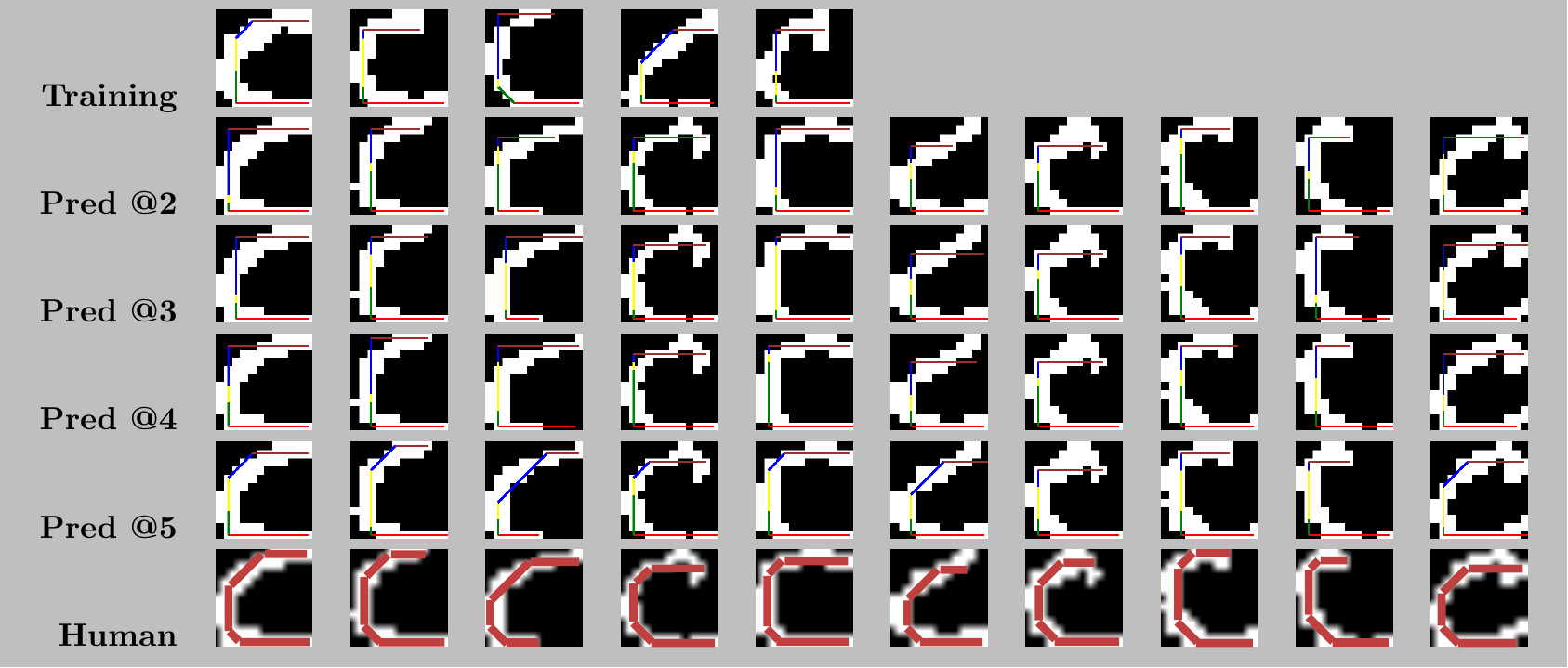}
	\caption{\label{fig:charC} Results for the ``C'' $12\times12$ character
drawing task.}
\end{figure}

\begin{figure}[H]
	\includegraphics{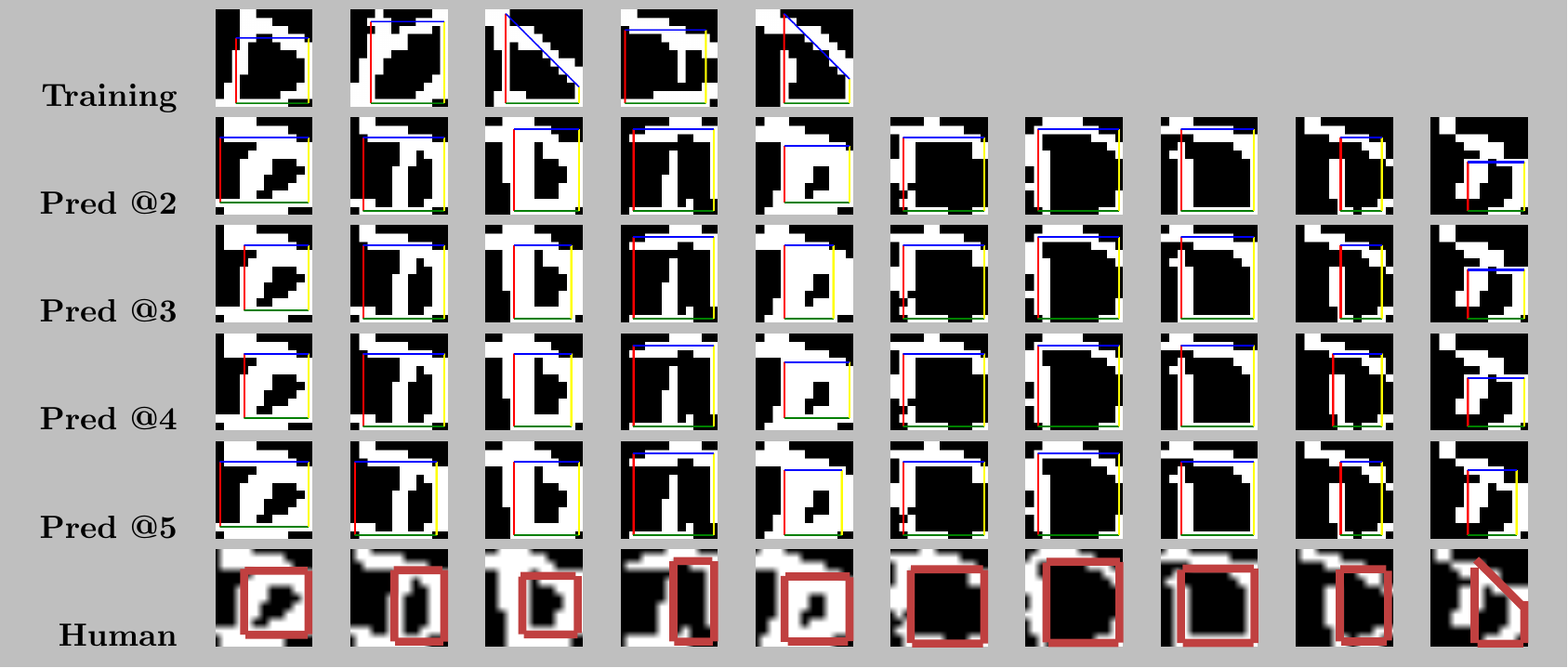}
	\caption{\label{fig:charD} Results for the ``D'' $12\times12$ character
drawing task.}
\end{figure}

\begin{figure}[H]
	\includegraphics{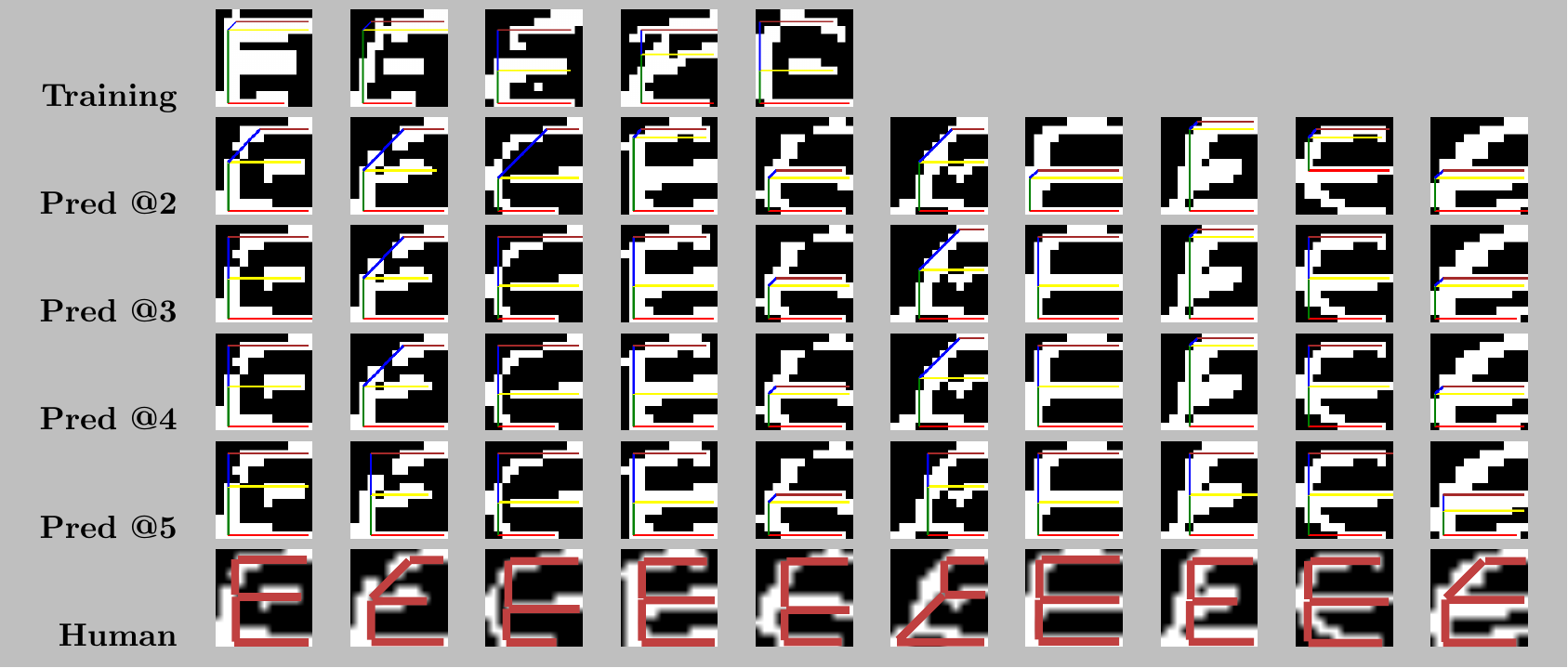}
	\caption{\label{fig:charE} Results for the ``E'' $12\times12$ character
drawing task.}
\end{figure}

We selected for each letter five images at random out of the 39
available to be employed as training instances. For each of these, we
used \optimathsat to generate a ``perfect'' vectorial representation
according to a human-provided letter template (similar to the
``looking like an A'' rule above), obtaining a training set of five fully
supervised images. Please note that the training outputs generated by this
process may not be optimal from a ``perceptual'' perspective, but only with
respect to the ``looking like an A'' rule.
The resulting supervision obtained with this procedure---which can be seen in
the first rows of Figures~\ref{fig:charA} to \ref{fig:charE}---is, in some
cases, very noisy, and depends crucially on the quality of the character image.
This is particularly relevant for the ``B'', which is the most geometrically
complex of the characters and thus more difficult to capture with bitmap
images.

For each letter, we randomly sampled a test set of 10 instances out of the 33
non-training images. Then we learned a model for each letter, and used it to
infer the vectorial representation of the test images.  In order to assess the
robustness of the learning method with respect to the amount of available
supervision, we repeated this procedure four times, each time adding a training
instance to the training set: the number of instances in the training set
grows from 2 (the first two training images) to 5 (all of the training images).
We indicate the predictions obtained by models learned with $k$ examples as
{\em pred@k}.
The number of segments $m$ was known during both training and inference. In
particular, we used 4 segments for the ``D'', 5 segments for the ``C'' and
``E'', 7 segments for the ``A'', and 9 segments for the ``B''.
The output for all letters can be found in Figures~\ref{fig:charA} to
\ref{fig:charE}, from the second to the fifth rows of each figure.

We sped up the computations in two ways. First, during learning we
imposed a 2 minute timeout on the separation oracle used for training.
Consequently most invocations to the separation routine did return an
approximate solution. Analyzing the weights, we found that this change had
little effect on the learned model (data not shown).
This can be explained by observing that the cutting-plane algorithm does not
actually {\em require} the separation oracle to be perfect: as long as the
approximation is consistent with the hard rules (which is necessarily the case
even in sub-optimal solutions), the constraint added to the working set
$\mathcal{W}$ at each iteration still restricts the quadratic program in a
sound manner (see Algorithm~\ref{alg:cp1slack}).  As a consequence, the
learning procedure is still guaranteed to find an $\epsilon$-approximate
solution, but it may take more iterations to converge.
This trick allows training to terminate very quickly (in the order of minutes
for a set of five training examples). Furthermore, it enables the user to
fine tune the trade-off between separation complexity and number of QP
sub-problems to be solved.

Second, prior to inference we convert the learned weights associated to the
segment and connection features into {\em hard} constraints and add them to the
learned model. This way we constrain \optimathsat to search for solutions that
do respect the learned weights, while still allowing for some flexibility in
the choice of the actual solution.
In practice, for each per-segment feature (i.e. those associated to {\tt
increasing}, {\tt decreasing} and {\tt right} soft constraints)
and connection features (i.e. {\tt h2t}, {\tt t2t}, {\em etc.}) with a positive
weight, we add the corresponding hard rule. If more than one weight is positive
for any given segment/connection, we add the disjuction of the hard rules to
the model.

\begin{figure}[H]
	\includegraphics[width=1\linewidth]{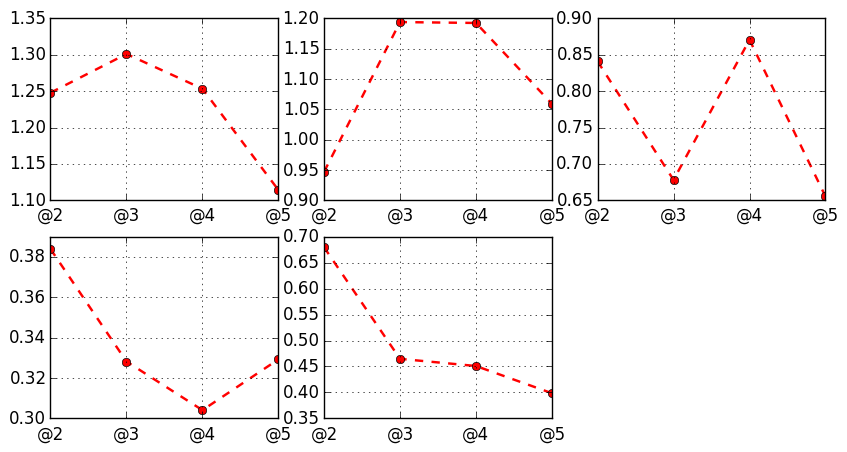}
	\caption{\label{fig:charOverlap} Average distance between the predicted
vectorial images and the human-made ones, while increasing the number of
training instances. The {\bf @k} ticks on the $x$-axis indicate the size of the
training set. The $y$-axis represents the sum of per-segment distances averaged
over all images in the test set. From left to right, top to bottom: results for
``A'', ``B'', ``C'', ``D'', and ``E''.}
\end{figure}

As a quantitative measure of the quality of the predictions, we also report the
distance between the generated vectorial representation $\lmty$ for each letter
and a corresponding human-made gold standard $\lmty'$. Here the error is
computed by first aligning the segments using an optimal translation, and
then summing the distances between all corresponding segment endpoints.
The human generated images respect the same ``looking like
an X'' rule used for generating the training set, i.e. they have the same
number of segments, drawn in the same order and within the same set of allowed
orientations. One minor difference is they do not follow the requirement that
segment endpoints match, for simplicity; this has little impact on the resulting
values.
The values in Figure~\ref{fig:charOverlap} are the average over all instances
in the test set, when varying the training set size.

The results show that LMT is able to address the character drawing problem and
produce reasonable outputs for all the target letters. Please note that both
the coordinates and the number of character pixels can vary widely between test
instances, and our results highlight the generalization ability of our method.
Furthermore, the predictions tend to get closer to the human-provided
segmentations as the number of training instances increases.

For the simplest cases (i.e. ``C'', ``D'', and ``E'', drawn using four to five
segments), the outcome is unambiguous: there are only a handful of cases for
``E'' where two horizontal segments are too close, and this is only due to the
fact that we do not encode (soft) constraints on the segment lenghts. The only
real issue is with the ``D'', where the fourth (blue) segment is never
predicted as diagonal, despite there being two examples with that property in
the training set. The main reason is that in none of the test images it is
possible to draw a $45^\circ$ diagonal without sacrificing pixel coverage. None
of the predictions looks perceptually ``wrong''.

More complex letters like the ``A'' and ``B'', with a higher number of
segments, also show a similar behavior: the generated letters indeed
generalize the given examples. However, there are a handful of
predictions that are sub-optimal with respect to coverage of the
bitmap character (e.g. see the first column in Figure~\ref{fig:charA})
or do not represent a ``perceptually'' correct character, a behaviour
which is less frequent for the larger training sets
(e.g. second-to-last row of Figure~\ref{fig:charA}). This can be
explained by (i) the complexity of the 7- and (especially) the
9-segment inference problems, (ii) the fact that more segments imply
more features, and consequently may require more examples to be
learned correctly, and (iii) the fact that our cost function does not
fully discriminate between perceptually different solutions.

The distance-to-human results in Figure~\ref{fig:charOverlap} also show how the
algorithm produces more perceptually reasonable predictions as the training set
increases: while the values do fluctuate, in all cases the
distance at {\em pred@5} is lower than that at {\em pred@2}.
%
Summarizing, excluding cases of pathologically bad inputs---such as the
third ``B'' training examples leading to the bad performance in the {\em pred@3}
and {\em pred@4} experiments---LMT is able to learn
an appropriate model for each letter and generalize the learned template over
unseen inputs.

In this paper we only consider the case where the training data is
labeled with a fully observed vectorial letter. Our method, however,
can in principle be extended to work with partially observed
supervision, e.g. with training instances labeled only with the
character type, in order to discover the vectorial representation.
More on this point can be found in the next section.

\section{Conclusions} 
\label{sec:conc}

In this work we presented a novel class of methods for structured
learning problems with mixed Boolean and numerical variables. These
methods, termed Learning Modulo Theories, are based on a combination
of structured-output SVMs and Satisfiability Modulo Theories. In
contrast to classical First-Order Logic, SMT allows to natively
describe, and reason over, numerical variables and mixed
logical/arithmetical constraints. By leveraging the existing
state-of-the-art OMT solvers, LMT is well-suited for dealing with
hybrid constructive learning problems, avoiding the combinatorial
explosion of the state space that would affect an equivalent FOL
formulation.

Experimental results on both artificial and real world datasets show
the potential of this approach. The stairway application is a simple
instance of layout problem, where the task is to find an optimal
layout subject to a set of constraints. Automated or interactive
layout synthesis has a broad range of potential applications,
including urban pattern layout~\cite{YanWanVouWon13}, decorative
mosaics~\cite{Hau2001} and furniture
arrangement~\cite{YuEtAl2011,MerEtAl2011,YehEtAl2012}. Note that many
spatial constraints can be encoded in terms of relationships between
blocks~\cite{YuEtAl2011}. Existing approaches typically design an
energy function to be minimized by stochastic search. Our approach
suggests how to automatically identify the relevant constraints and
their respective weights, and can accomodate hard constraints and
exact search. This is especially relevant for {\em water-tight}
layouts~\cite{PenYanWon14}, where the whole space needs to be filled
(i.e. no gaps or overlaps) by deforming basic elements from a
predetermined set of templates (as in residential building
layout~\cite{MerSchKol10}). The character drawing application shows
how to learn symbolic representations from a handful of very noisy
training instances.  Deep generative networks have been previously
used for similar tasks, see for instance the work by
Hinton~\cite{Hin2007} on generating images of digits with Deep
Boltzmann Machines. However, these methods do not learn symbolic
representations for characters and generate bitmaps rather than
vectorial representations. Furthermore, they require thousands of
training examples to be learned. Recent extensions have been developed
addressing the problem of learning from few~\cite{SalTenTor13} or even
single~\cite{LakSalTen13} examples, but they focus on clean images of
the target symbols. Generally speaking, the LMT framework allows to
introduce a learning stage in all application domains where SMT and
OMT approaches have shown their potential, ranging, e.g., from
hardware and software verification
\cite{beyercgks09,fcnss_fmcad10,MouraB11}, to engineering of chemical
reactions~\cite{FagFlaMerPet12} and synthetic
biology~\cite{YorWinHamKug13}.

This work can be extended in a number of directions. First, the
current formulation assumes knowledge of the desired output for
training examples. This requirement can be loosened by introducting
latent variables for the unobserved part of the output, to be
maximized over during training~\cite{YuJoa09}. Second, OMT is
currently limited to quantifier free formulae and linear algebra for
what concerns numeric theories. The former requires to ground all
predicates before performing inference, while the latter prevents the
formulation of non-linear constraints, e.g. on areas and Euclidean
distances. Some attempts to extend SMT solvers to both quantified
formulae~\cite{Rum08,BauTin11,Kruglov13} and non-linear
arithmetic~\cite{FranzleHTRS07,JovanovicM12} have been presented in the
literature; although the state of the art of these estensions is not
satisfactory yet, we can rather easily extend our framework in these
directions as soon as the underlying SMT technology gets mature
enough. Finally, LMT is currently focused on the task of finding the
maximal configuration, and cannot compute marginal probabilities. We
are planning to introduce support for probability computation by
leveraging ideas from weighted model counting~\cite{ChaDar08}.

\section*{Acknowledgements}

We would like to thank Luc De Raedt, Guy Van den Broeck and Bernd
Gutmann for useful discussions.











\bibliographystyle{model1-num-names}

\bibliography{paper.bib}

\begin{thebibliography}{67}
\expandafter\ifx\csname natexlab\endcsname\relax\def\natexlab#1{#1}\fi
\providecommand{\bibinfo}[2]{#2}
\ifx\xfnm\relax \def\xfnm[#1]{\unskip,\space#1}\fi
\bibitem[{Getoor and Taskar(2007)}]{getoor2007introduction}
\bibinfo{author}{L.~Getoor}, \bibinfo{author}{B.~Taskar},
  \bibinfo{title}{Introduction to statistical relational learning},
  \bibinfo{publisher}{The MIT press}, \bibinfo{year}{2007}.
\bibitem[{Bakir et~al.(2007)Bakir, Hofmann, Sch\"{o}lkopf, Smola, Taskar, and
  Vishwanathan}]{BakHofSchSmoTasVis07}
\bibinfo{author}{G.~H. Bakir}, \bibinfo{author}{T.~Hofmann},
  \bibinfo{author}{B.~Sch\"{o}lkopf}, \bibinfo{author}{A.~J. Smola},
  \bibinfo{author}{B.~Taskar}, \bibinfo{author}{S.~V.~N. Vishwanathan},
  \bibinfo{title}{Predicting Structured Data (Neural Information Processing)},
  \bibinfo{publisher}{The MIT Press}, \bibinfo{year}{2007}.
\bibitem[{Lippi and Frasconi(2009)}]{Lip09}
\bibinfo{author}{M.~Lippi}, \bibinfo{author}{P.~Frasconi},
\newblock \bibinfo{title}{Prediction of protein β-residue contacts by markov
  logic networks with grounding-specific weights},
\newblock \bibinfo{journal}{Bioinformatics} \bibinfo{volume}{25}
  (\bibinfo{year}{2009}) \bibinfo{pages}{2326--2333}.
\bibitem[{Broecheler et~al.(2010)Broecheler, Mihalkova, and
  Getoor}]{BroMihGet10}
\bibinfo{author}{M.~Broecheler}, \bibinfo{author}{L.~Mihalkova},
  \bibinfo{author}{L.~Getoor},
\newblock \bibinfo{title}{Probabilistic similarity logic},
\newblock in: \bibinfo{booktitle}{Uncertainty in Artificial Intelligence
  (UAI)}, \bibinfo{year}{2010}, pp. \bibinfo{pages}{73--82}.
\bibitem[{Ku$\check{z}$elka et~al.(2011)Ku$\check{z}$elka, Szab{\'o}v{\'a},
  Holec, and $\check{Z}$elezn{\'y}}]{Kuz11}
\bibinfo{author}{O.~Ku$\check{z}$elka}, \bibinfo{author}{A.~Szab{\'o}v{\'a}},
  \bibinfo{author}{M.~Holec}, \bibinfo{author}{F.~$\check{Z}$elezn{\'y}},
\newblock \bibinfo{title}{Gaussian logic for predictive classification},
\newblock in: \bibinfo{editor}{D.~Gunopulos}, \bibinfo{editor}{T.~Hofmann},
  \bibinfo{editor}{D.~Malerba}, \bibinfo{editor}{M.~Vazirgiannis} (Eds.),
  \bibinfo{booktitle}{Machine Learning and Knowledge Discovery in Databases},
  volume \bibinfo{volume}{6912} of \textit{\bibinfo{series}{Lecture Notes in
  Computer Science}}, \bibinfo{publisher}{Springer Berlin Heidelberg},
  \bibinfo{year}{2011}, pp. \bibinfo{pages}{277--292}.
\bibitem[{Diligenti et~al.(2012)Diligenti, Gori, Maggini, and Rigutini}]{Dil12}
\bibinfo{author}{M.~Diligenti}, \bibinfo{author}{M.~Gori},
  \bibinfo{author}{M.~Maggini}, \bibinfo{author}{L.~Rigutini},
\newblock \bibinfo{title}{Bridging logic and kernel machines},
\newblock \bibinfo{journal}{Machine Learning} \bibinfo{volume}{86}
  (\bibinfo{year}{2012}) \bibinfo{pages}{57--88}.
\bibitem[{Goodman et~al.(2008)Goodman, Mansinghka, Roy, Bonawitz, and
  Tenenbaum}]{church08}
\bibinfo{author}{N.~D. Goodman}, \bibinfo{author}{V.~K. Mansinghka},
  \bibinfo{author}{D.~M. Roy}, \bibinfo{author}{K.~Bonawitz},
  \bibinfo{author}{J.~B. Tenenbaum},
\newblock \bibinfo{title}{Church: a language for generative models},
\newblock in: \bibinfo{editor}{D.~A. McAllester},
  \bibinfo{editor}{P.~Myllym{\"a}ki} (Eds.), \bibinfo{booktitle}{UAI},
  \bibinfo{publisher}{AUAI Press}, \bibinfo{year}{2008}, pp.
  \bibinfo{pages}{220--229}.
\bibitem[{Wang and Domingos(2008)}]{Wan08}
\bibinfo{author}{J.~Wang}, \bibinfo{author}{P.~Domingos},
\newblock \bibinfo{title}{Hybrid markov logic networks},
\newblock in: \bibinfo{booktitle}{Proceedings of the 23rd National Conference
  on Artificial Intelligence - Volume 2}, AAAI'08, \bibinfo{publisher}{AAAI
  Press}, \bibinfo{year}{2008}, pp. \bibinfo{pages}{1106--1111}.
\bibitem[{Närman et~al.(2010)Närman, Buschle, König, and
  Johnson}]{NarBusKon10}
\bibinfo{author}{P.~Närman}, \bibinfo{author}{M.~Buschle},
  \bibinfo{author}{J.~König}, \bibinfo{author}{P.~Johnson},
\newblock \bibinfo{title}{Hybrid probabilistic relational models for system
  quality analysis.},
\newblock in: \bibinfo{booktitle}{EDOC}, \bibinfo{publisher}{IEEE Computer
  Society}, \bibinfo{year}{2010}, pp. \bibinfo{pages}{57--66}.
\bibitem[{Gutmann et~al.(2011)Gutmann, Jaeger, and De~Raedt}]{Gut11}
\bibinfo{author}{B.~Gutmann}, \bibinfo{author}{M.~Jaeger},
  \bibinfo{author}{L.~De~Raedt},
\newblock \bibinfo{title}{Extending problog with continuous distributions},
\newblock in: \bibinfo{editor}{P.~Frasconi}, \bibinfo{editor}{F.~Lisi} (Eds.),
  \bibinfo{booktitle}{Inductive Logic Programming}, volume
  \bibinfo{volume}{6489} of \textit{\bibinfo{series}{Lecture Notes in Computer
  Science}}, \bibinfo{publisher}{Springer Berlin Heidelberg},
  \bibinfo{year}{2011}, pp. \bibinfo{pages}{76--91}.
\bibitem[{Choi and Amir(2012)}]{ChoAmi12}
\bibinfo{author}{J.~Choi}, \bibinfo{author}{E.~Amir},
\newblock \bibinfo{title}{Lifted relational variational inference},
\newblock in: \bibinfo{editor}{N.~de~Freitas}, \bibinfo{editor}{K.~P. Murphy}
  (Eds.), \bibinfo{booktitle}{UAI'12: Proceedings of the Twenty-Eigth
  Conference on Uncertainty in Artificial Intelligence},
  \bibinfo{publisher}{AUAI Press}, \bibinfo{year}{2012}, pp.
  \bibinfo{pages}{196--206}.
\bibitem[{Islam et~al.(2012)Islam, Ramakrishnan, and
  Ramakrishnan}]{IslRamRam2012}
\bibinfo{author}{M.~a. Islam}, \bibinfo{author}{C.~r. Ramakrishnan},
  \bibinfo{author}{I.~v. Ramakrishnan},
\newblock \bibinfo{title}{Inference in probabilistic logic programs with
  continuous random variables},
\newblock \bibinfo{journal}{Theory Pract. Log. Program.} \bibinfo{volume}{12}
  (\bibinfo{year}{2012}) \bibinfo{pages}{505--523}.
\bibitem[{Barrett et~al.(2009)Barrett, Sebastiani, Seshia, and
  Tinelli}]{BSST09HBSAT}
\bibinfo{author}{C.~Barrett}, \bibinfo{author}{R.~Sebastiani},
  \bibinfo{author}{S.~A. Seshia}, \bibinfo{author}{C.~Tinelli},
  \bibinfo{title}{Satisfiability Modulo Theories}, in:
  \cite{HandbookOfSAT2009}, pp. \bibinfo{pages}{825--885}.
\bibitem[{Nieuwenhuis and Oliveras(2006)}]{nieuwenhuis_sat06}
\bibinfo{author}{R.~Nieuwenhuis}, \bibinfo{author}{A.~Oliveras},
\newblock \bibinfo{title}{{On SAT Modulo Theories and Optimization Problems}},
\newblock in: \bibinfo{booktitle}{Proc. Theory and Applications of
  Satisfiability Testing - SAT 2006}, volume \bibinfo{volume}{4121} of
  \textit{\bibinfo{series}{LNCS}}, \bibinfo{publisher}{Springer},
  \bibinfo{year}{2006}.
\bibitem[{Cimatti et~al.(2010)Cimatti, Franz{\'e}n, Griggio, Sebastiani, and
  Stenico}]{cimattifgss10}
\bibinfo{author}{A.~Cimatti}, \bibinfo{author}{A.~Franz{\'e}n},
  \bibinfo{author}{A.~Griggio}, \bibinfo{author}{R.~Sebastiani},
  \bibinfo{author}{C.~Stenico},
\newblock \bibinfo{title}{Satisfiability modulo the theory of costs:
  Foundations and applications},
\newblock in: \bibinfo{booktitle}{Proc. Tools and Algorithms for the
  Construction and Analysis of Systems, TACAS}, volume \bibinfo{volume}{6015}
  of \textit{\bibinfo{series}{LNCS}}, \bibinfo{publisher}{Springer},
  \bibinfo{year}{2010}, pp. \bibinfo{pages}{99--113}.
\bibitem[{Cimatti et~al.(2013)Cimatti, Griggio, Schaafsma, and
  Sebastiani}]{cgss_sat13_maxsmt}
\bibinfo{author}{A.~Cimatti}, \bibinfo{author}{A.~Griggio},
  \bibinfo{author}{B.~J. Schaafsma}, \bibinfo{author}{R.~Sebastiani},
\newblock \bibinfo{title}{{A Modular Approach to MaxSAT Modulo Theories}},
\newblock in: \bibinfo{booktitle}{International Conference on Theory and
  Applications of Satisfiability Testing, SAT}, volume \bibinfo{volume}{7962}
  of \textit{\bibinfo{series}{LNCS}}, \bibinfo{publisher}{Springer},
  \bibinfo{year}{2013}.
\bibitem[{Li and Many{\`a}(2009)}]{LM09HBSAT}
\bibinfo{author}{C.~M. Li}, \bibinfo{author}{F.~Many{\`a}},
  \bibinfo{title}{MaxSAT, Hard and Soft Constraints}, in:
  \cite{HandbookOfSAT2009}, pp. \bibinfo{pages}{613--631}.
\bibitem[{Sebastiani and Tomasi(2012)}]{st-ijcar12}
\bibinfo{author}{R.~Sebastiani}, \bibinfo{author}{S.~Tomasi},
\newblock \bibinfo{title}{{Optimization in SMT with LA(Q) Cost Functions}},
\newblock in: \bibinfo{booktitle}{proc. International Joint Conference on
  Automated Reasoning, IJCAR}, volume \bibinfo{volume}{7364} of
  \textit{\bibinfo{series}{LNAI}}, \bibinfo{publisher}{Springer},
  \bibinfo{year}{2012}, pp. \bibinfo{pages}{484--498}.
\bibitem[{Sebastiani and Tomasi(2014)}]{sebastiani14_optimathsat}
\bibinfo{author}{R.~Sebastiani}, \bibinfo{author}{S.~Tomasi},
\newblock \bibinfo{title}{{Optimization Modulo Theories with Linear Rational
  Costs}},
\newblock \bibinfo{journal}{ACM Transactions on Computational Logics}
  (\bibinfo{year}{2014}). \bibinfo{note}{To appear. Available as
  \url{http://arxiv.org/pdf/1410.6039.pdf}.}
\bibitem[{Li et~al.(2014)Li, Albarghouthi, Kincad, Gurfinkel, and
  Chechik}]{li_popl14}
\bibinfo{author}{Y.~Li}, \bibinfo{author}{A.~Albarghouthi},
  \bibinfo{author}{Z.~Kincad}, \bibinfo{author}{A.~Gurfinkel},
  \bibinfo{author}{M.~Chechik},
\newblock \bibinfo{title}{{Symbolic Optimization with SMT Solvers}},
\newblock in: \bibinfo{booktitle}{Proceedings of the 41st ACM SIGPLAN-SIGACT
  Symposium on Principles of Programming Languages}, POPL '14,
  \bibinfo{publisher}{ACM}, \bibinfo{address}{New York, NY, USA},
  \bibinfo{year}{2014}, pp. \bibinfo{pages}{607--618}.
\bibitem[{Tsochantaridis et~al.(2005)Tsochantaridis, Joachims, Hofmann, and
  Altun}]{tsochantaridis2005large}
\bibinfo{author}{I.~Tsochantaridis}, \bibinfo{author}{T.~Joachims},
  \bibinfo{author}{T.~Hofmann}, \bibinfo{author}{Y.~Altun},
\newblock \bibinfo{title}{Large margin methods for structured and
  interdependent output variables},
\newblock \bibinfo{journal}{J. Mach. Learn. Res.} \bibinfo{volume}{6}
  (\bibinfo{year}{2005}) \bibinfo{pages}{1453--1484}.
\bibitem[{Joachims et~al.(2009)Joachims, Hofmann, Yue, and
  Yu}]{joachims2009predicting}
\bibinfo{author}{T.~Joachims}, \bibinfo{author}{T.~Hofmann},
  \bibinfo{author}{Y.~Yue}, \bibinfo{author}{C.-N. Yu},
\newblock \bibinfo{title}{Predicting structured objects with support vector
  machines},
\newblock \bibinfo{journal}{Communications of the ACM} \bibinfo{volume}{52}
  (\bibinfo{year}{2009}) \bibinfo{pages}{97--104}.
\bibitem[{Campigotto et~al.(2011)Campigotto, Passerini, and
  Battiti}]{stochLogicUtFun2010}
\bibinfo{author}{P.~Campigotto}, \bibinfo{author}{A.~Passerini},
  \bibinfo{author}{R.~Battiti},
\newblock \bibinfo{title}{{Active Learning of Combinatorial Features for
  Interactive Optimization}},
\newblock in: \bibinfo{booktitle}{Proceedings of the 5th Learning and
  Intelligent OptimizatioN Conference (LION V), Rome, Italy, Jan 17-21, 2011},
  LNCS, \bibinfo{publisher}{Springer Verlag}, \bibinfo{year}{2011}.
\bibitem[{Choi et~al.(2010)Choi, Hill, and Amir}]{ChoHilAmi10}
\bibinfo{author}{J.~Choi}, \bibinfo{author}{D.~Hill},
  \bibinfo{author}{E.~Amir},
\newblock \bibinfo{title}{Lifted inference for relational continuous models},
\newblock in: \bibinfo{booktitle}{UAI'10: Proceedings of the Twenty-Sixth
  Conference on Uncertainty in Artificial Intelligence}.
\bibitem[{Choi et~al.(2011)Choi, Guzmán-Rivera, and Amir}]{ChoGuzAmi11}
\bibinfo{author}{J.~Choi}, \bibinfo{author}{A.~Guzmán-Rivera},
  \bibinfo{author}{E.~Amir},
\newblock \bibinfo{title}{Lifted relational kalman filtering.},
\newblock in: \bibinfo{booktitle}{Proceedings of IJCAI'11}, pp.
  \bibinfo{pages}{2092--2099}.
\bibitem[{Ahmadi et~al.(2011)Ahmadi, Kersting, and Sanner}]{AhmKerSan11}
\bibinfo{author}{B.~Ahmadi}, \bibinfo{author}{K.~Kersting},
  \bibinfo{author}{S.~Sanner},
\newblock \bibinfo{title}{Multi-evidence lifted message passing, with
  application to pagerank and the kalman filter},
\newblock in: \bibinfo{booktitle}{Proceedings of IJCAI'11}, pp.
  \bibinfo{pages}{1152--1158}.
\bibitem[{Murphy(1998)}]{Murphy98hybrid}
\bibinfo{author}{K.~P. Murphy}, \bibinfo{title}{Inference and Learning in
  Hybrid Bayesian Networks}, \bibinfo{type}{Technical Report}, University of
  California, Berkeley, Computer Science Division, \bibinfo{year}{1998}.
\bibitem[{Lauritzen(1992)}]{HybridBN}
\bibinfo{author}{S.~L. Lauritzen},
\newblock \bibinfo{title}{Propagation of probabilities, means, and variances in
  mixed graphical association models},
\newblock \bibinfo{journal}{Journal of the American Statistical Association}
  \bibinfo{volume}{87} (\bibinfo{year}{1992}) \bibinfo{pages}{1098--1108}.
\bibitem[{Sato(1995)}]{Sato95}
\bibinfo{author}{T.~Sato},
\newblock \bibinfo{title}{A statistical learning method for logic programs with
  distribution semantics.},
\newblock in: \bibinfo{editor}{L.~Sterling} (Ed.), \bibinfo{booktitle}{ICLP},
  \bibinfo{publisher}{MIT Press}, \bibinfo{year}{1995}, pp.
  \bibinfo{pages}{715--729}.
\bibitem[{Islam(2012)}]{IslamPhd12}
\bibinfo{author}{M.~A. Islam}, \bibinfo{title}{Inference and Learning in
  Probabilistic Logic Programs with Continuous Random Variables}, Ph.D. thesis,
  Stony Brook University, \bibinfo{year}{2012}.
\bibitem[{Griggio et~al.(2011)Griggio, Phan, Sebastiani, and
  Tomasi}]{frocos11_walksmt}
\bibinfo{author}{A.~Griggio}, \bibinfo{author}{Q.~S. Phan},
  \bibinfo{author}{R.~Sebastiani}, \bibinfo{author}{S.~Tomasi},
\newblock \bibinfo{title}{{Stochastic Local Search for SMT: Combining Theory
  Solvers with WalkSAT}},
\newblock in: \bibinfo{booktitle}{Frontiers of Combining Systems, FroCoS'11},
  volume \bibinfo{volume}{6989} of \textit{\bibinfo{series}{LNAI}},
  \bibinfo{publisher}{Springer}, \bibinfo{year}{2011}.
\bibitem[{De~Raedt et~al.(2007)De~Raedt, Kimmig, and Toivonen}]{problog07}
\bibinfo{author}{L.~De~Raedt}, \bibinfo{author}{A.~Kimmig},
  \bibinfo{author}{H.~Toivonen},
\newblock \bibinfo{title}{Problog: A probabilistic prolog and its application
  in link discovery.},
\newblock in: \bibinfo{editor}{M.~M. Veloso} (Ed.), \bibinfo{booktitle}{IJCAI},
  \bibinfo{year}{2007}, pp. \bibinfo{pages}{2462--2467}.
\bibitem[{Gutmann et~al.(2011)Gutmann, Thon, Kimmig, Bruynooghe, and
  De~Raedt}]{GutThoKim11}
\bibinfo{author}{B.~Gutmann}, \bibinfo{author}{I.~Thon},
  \bibinfo{author}{A.~Kimmig}, \bibinfo{author}{M.~Bruynooghe},
  \bibinfo{author}{L.~De~Raedt},
\newblock \bibinfo{title}{The magic of logical inference in probabilistic
  programming.},
\newblock \bibinfo{journal}{TPLP} \bibinfo{volume}{11} (\bibinfo{year}{2011})
  \bibinfo{pages}{663--680}.
\bibitem[{Prestwich(2009)}]{Pre09HBSAT}
\bibinfo{author}{S.~Prestwich}, \bibinfo{title}{CNF Encodings}, in:
  \cite{HandbookOfSAT2009}, pp. \bibinfo{pages}{75--97}.
\bibitem[{Marques-Silva et~al.(2009)Marques-Silva, Lynce, and
  Malik}]{MSLM09HBSAT}
\bibinfo{author}{J.~P. Marques-Silva}, \bibinfo{author}{I.~Lynce},
  \bibinfo{author}{S.~Malik}, \bibinfo{title}{Conflict-Driven Clause Learning
  SAT Solvers}, in:  \cite{HandbookOfSAT2009}, pp. \bibinfo{pages}{131--153}.
\bibitem[{Biere et~al.(2009)Biere, Heule, van Maaren, and
  Walsh}]{HandbookOfSAT2009}
\bibinfo{editor}{A.~Biere}, \bibinfo{editor}{M.~J.~H. Heule},
  \bibinfo{editor}{H.~van Maaren}, \bibinfo{editor}{T.~Walsh} (Eds.),
  \bibinfo{title}{Handbook of Satisfiability}, Frontiers in Artificial
  Intelligence and Applications, \bibinfo{publisher}{IOS Press},
  \bibinfo{year}{2009}.
\bibitem[{de~Moura and Bj{\o}rner(2011)}]{MouraB11}
\bibinfo{author}{L.~de~Moura}, \bibinfo{author}{N.~Bj{\o}rner},
\newblock \bibinfo{title}{Satisfiability modulo theories: introduction and
  applications},
\newblock \bibinfo{journal}{Commun. ACM} \bibinfo{volume}{54}
  (\bibinfo{year}{2011}) \bibinfo{pages}{69--77}.
\bibitem[{Sebastiani(2007)}]{sebastiani07}
\bibinfo{author}{R.~Sebastiani},
\newblock \bibinfo{title}{{Lazy Satisfiability Modulo Theories}},
\newblock \bibinfo{journal}{Journal on Satisfiability, Boolean Modeling and
  Computation, JSAT} \bibinfo{volume}{3} (\bibinfo{year}{2007})
  \bibinfo{pages}{141--224}.
\bibitem[{Balas(2010)}]{balasdisjunctive}
\bibinfo{author}{E.~Balas},
\newblock \bibinfo{title}{Disjunctive programming},
\newblock in: \bibinfo{editor}{M.~Junger}, \bibinfo{editor}{T.~M. Liebling},
  \bibinfo{editor}{D.~Naddef}, \bibinfo{editor}{G.~L. Nemhauser},
  \bibinfo{editor}{W.~R. Pulleyblank}, \bibinfo{editor}{G.~Reinelt},
  \bibinfo{editor}{G.~Rinaldi}, \bibinfo{editor}{L.~A. Wolsey} (Eds.),
  \bibinfo{booktitle}{50 Years of Integer Programming 1958-2008},
  \bibinfo{publisher}{Springer Berlin Heidelberg}, \bibinfo{year}{2010}, pp.
  \bibinfo{pages}{283--340}.
\bibitem[{Lodi(2009)}]{lodi}
\bibinfo{author}{A.~Lodi},
\newblock \bibinfo{title}{{Mixed Integer Programming Computation}},
\newblock in: \bibinfo{booktitle}{50 Years of Integer Programming 1958-2008},
  \bibinfo{publisher}{Springer-Verlag}, \bibinfo{year}{2009}, pp.
  \bibinfo{pages}{619--645}.
\bibitem[{Sawaya and Grossmann(2012)}]{sawayagrossmann2012}
\bibinfo{author}{N.~W. Sawaya}, \bibinfo{author}{I.~E. Grossmann},
\newblock \bibinfo{title}{A hierarchy of relaxations for linear generalized
  disjunctive programming},
\newblock \bibinfo{journal}{European Journal of Operational Research}
  \bibinfo{volume}{216} (\bibinfo{year}{2012}) \bibinfo{pages}{70--82}.
\bibitem[{Jaffar and Maher(1994)}]{Jaffar94}
\bibinfo{author}{J.~Jaffar}, \bibinfo{author}{M.~J. Maher},
\newblock \bibinfo{title}{{Constraint Logic Programming: A Survey}},
\newblock \bibinfo{journal}{Journal of Logic Programming}
  \bibinfo{volume}{19/20} (\bibinfo{year}{1994}) \bibinfo{pages}{503--581}.
\bibitem[{Codognet and Diaz(1996)}]{CodDia96}
\bibinfo{author}{P.~Codognet}, \bibinfo{author}{D.~Diaz},
\newblock \bibinfo{title}{Compiling constraints in clp(fd)},
\newblock \bibinfo{journal}{The Journal of Logic Programming}
  \bibinfo{volume}{27} (\bibinfo{year}{1996}) \bibinfo{pages}{185 -- 226}.
\bibitem[{Jaffar et~al.(1992)Jaffar, Michaylov, Stuckey, and Yap}]{JafJox92}
\bibinfo{author}{J.~Jaffar}, \bibinfo{author}{S.~Michaylov},
  \bibinfo{author}{P.~J. Stuckey}, \bibinfo{author}{R.~H.~C. Yap},
\newblock \bibinfo{title}{The clp( r ) language and system},
\newblock \bibinfo{journal}{ACM Trans. Program. Lang. Syst.}
  \bibinfo{volume}{14} (\bibinfo{year}{1992}) \bibinfo{pages}{339--395}.
\bibitem[{Cimatti et~al.(2013)Cimatti, Griggio, Schaafsma, and
  Sebastiani}]{mathsat5_tacas13}
\bibinfo{author}{A.~Cimatti}, \bibinfo{author}{A.~Griggio},
  \bibinfo{author}{B.~J. Schaafsma}, \bibinfo{author}{R.~Sebastiani},
\newblock \bibinfo{title}{{The MathSAT 5 SMT Solver}},
\newblock in: \bibinfo{booktitle}{Tools and Algorithms for the Construction and
  Analysis of Systems, TACAS'13.}, volume \bibinfo{volume}{7795} of
  \textit{\bibinfo{series}{LNCS}}, \bibinfo{publisher}{Springer},
  \bibinfo{year}{2013}, pp. \bibinfo{pages}{95--109}.
\bibitem[{R\"ummer(2008)}]{Rum08}
\bibinfo{author}{P.~R\"ummer},
\newblock \bibinfo{title}{A constraint sequent calculus for first-order logic
  with linear integer arithmetic},
\newblock in: \bibinfo{editor}{I.~Cervesato}, \bibinfo{editor}{H.~Veith},
  \bibinfo{editor}{A.~Voronkov} (Eds.), \bibinfo{booktitle}{Logic for
  Programming, Artificial Intelligence, and Reasoning}, volume
  \bibinfo{volume}{5330} of \textit{\bibinfo{series}{Lecture Notes in Computer
  Science}}, \bibinfo{publisher}{Springer Berlin Heidelberg},
  \bibinfo{year}{2008}, pp. \bibinfo{pages}{274--289}.
\bibitem[{Baumgartner and Tinelli(2011)}]{BauTin11}
\bibinfo{author}{P.~Baumgartner}, \bibinfo{author}{C.~Tinelli},
\newblock \bibinfo{title}{Model evolution with equality modulo built-in
  theories},
\newblock in: \bibinfo{editor}{N.~Bj{\o}rner},
  \bibinfo{editor}{V.~Sofronie-Stokkermans} (Eds.),
  \bibinfo{booktitle}{Automated Deduction – CADE-23}, volume
  \bibinfo{volume}{6803} of \textit{\bibinfo{series}{Lecture Notes in Computer
  Science}}, \bibinfo{publisher}{Springer Berlin Heidelberg},
  \bibinfo{year}{2011}, pp. \bibinfo{pages}{85--100}.
\bibitem[{Kruglov(2013)}]{Kruglov13}
\bibinfo{author}{E.~Kruglov}, \bibinfo{title}{Superposition modulo theory},
  Ph.D. thesis, Universit\"at des Saarlandes, \bibinfo{address}{Postfach
  151141, 66041 Saarbr\"ucken}, \bibinfo{year}{2013}.
\bibitem[{Joachims et~al.(2009)Joachims, Finley, and Yu}]{joachims2009cutting}
\bibinfo{author}{T.~Joachims}, \bibinfo{author}{T.~Finley},
  \bibinfo{author}{C.-N.~J. Yu},
\newblock \bibinfo{title}{Cutting-plane training of structural svms},
\newblock \bibinfo{journal}{Machine Learning} \bibinfo{volume}{77}
  (\bibinfo{year}{2009}) \bibinfo{pages}{27--59}.
\bibitem[{Yang et~al.(2013)Yang, Wang, Vouga, and Wonka}]{YanWanVouWon13}
\bibinfo{author}{Y.-L. Yang}, \bibinfo{author}{J.~Wang},
  \bibinfo{author}{E.~Vouga}, \bibinfo{author}{P.~Wonka},
\newblock \bibinfo{title}{Urban pattern: Layout design by hierarchical domain
  splitting},
\newblock \bibinfo{journal}{ACM Trans. Graph.} \bibinfo{volume}{32}
  (\bibinfo{year}{2013}) \bibinfo{pages}{181:1--181:12}.
\bibitem[{Hausner(2001)}]{Hau2001}
\bibinfo{author}{A.~Hausner},
\newblock \bibinfo{title}{Simulating decorative mosaics},
\newblock in: \bibinfo{booktitle}{Proceedings of the 28th Annual Conference on
  Computer Graphics and Interactive Techniques}, SIGGRAPH '01,
  \bibinfo{publisher}{ACM}, \bibinfo{address}{New York, NY, USA},
  \bibinfo{year}{2001}, pp. \bibinfo{pages}{573--580}.
\bibitem[{Yu et~al.(2011)Yu, Yeung, Tang, Terzopoulos, Chan, and
  Osher}]{YuEtAl2011}
\bibinfo{author}{L.-F. Yu}, \bibinfo{author}{S.-K. Yeung},
  \bibinfo{author}{C.-K. Tang}, \bibinfo{author}{D.~Terzopoulos},
  \bibinfo{author}{T.~F. Chan}, \bibinfo{author}{S.~J. Osher},
\newblock \bibinfo{title}{Make it home: Automatic optimization of furniture
  arrangement},
\newblock \bibinfo{journal}{ACM Trans. Graph.} \bibinfo{volume}{30}
  (\bibinfo{year}{2011}) \bibinfo{pages}{86:1--86:12}.
\bibitem[{Merrell et~al.(2011)Merrell, Schkufza, Li, Agrawala, and
  Koltun}]{MerEtAl2011}
\bibinfo{author}{P.~Merrell}, \bibinfo{author}{E.~Schkufza},
  \bibinfo{author}{Z.~Li}, \bibinfo{author}{M.~Agrawala},
  \bibinfo{author}{V.~Koltun},
\newblock \bibinfo{title}{Interactive furniture layout using interior design
  guidelines},
\newblock \bibinfo{journal}{ACM Trans. Graph.} \bibinfo{volume}{30}
  (\bibinfo{year}{2011}) \bibinfo{pages}{87:1--87:10}.
\bibitem[{Yeh et~al.(2012)Yeh, Yang, Watson, Goodman, and
  Hanrahan}]{YehEtAl2012}
\bibinfo{author}{Y.-T. Yeh}, \bibinfo{author}{L.~Yang},
  \bibinfo{author}{M.~Watson}, \bibinfo{author}{N.~D. Goodman},
  \bibinfo{author}{P.~Hanrahan},
\newblock \bibinfo{title}{Synthesizing open worlds with constraints using
  locally annealed reversible jump mcmc},
\newblock \bibinfo{journal}{ACM Trans. Graph.} \bibinfo{volume}{31}
  (\bibinfo{year}{2012}) \bibinfo{pages}{56:1--56:11}.
\bibitem[{Peng et~al.(2014)Peng, Yang, and Wonka}]{PenYanWon14}
\bibinfo{author}{C.-H. Peng}, \bibinfo{author}{Y.-L. Yang},
  \bibinfo{author}{P.~Wonka},
\newblock \bibinfo{title}{Computing layouts with deformable templates},
\newblock \bibinfo{journal}{ACM Trans. Graph.} \bibinfo{volume}{33}
  (\bibinfo{year}{2014}) \bibinfo{pages}{99:1--99:11}.
\bibitem[{Merrell et~al.(2010)Merrell, Schkufza, and Koltun}]{MerSchKol10}
\bibinfo{author}{P.~Merrell}, \bibinfo{author}{E.~Schkufza},
  \bibinfo{author}{V.~Koltun},
\newblock \bibinfo{title}{Computer-generated residential building layouts},
\newblock \bibinfo{journal}{ACM Trans. Graph.} \bibinfo{volume}{29}
  (\bibinfo{year}{2010}) \bibinfo{pages}{181:1--181:12}.
\bibitem[{Hinton(2007)}]{Hin2007}
\bibinfo{author}{G.~E. Hinton},
\newblock \bibinfo{title}{To recognize shapes first learn to generate images},
\newblock in: \bibinfo{booktitle}{Computational Neuroscience: Theoretical
  Insights into Brain Function}, \bibinfo{publisher}{Elsevier},
  \bibinfo{year}{2007}.
\bibitem[{Salakhutdinov et~al.(2013)Salakhutdinov, Tenenbaum, and
  Torralba}]{SalTenTor13}
\bibinfo{author}{R.~Salakhutdinov}, \bibinfo{author}{J.~B. Tenenbaum},
  \bibinfo{author}{A.~Torralba},
\newblock \bibinfo{title}{Learning with hierarchical-deep models.},
\newblock \bibinfo{journal}{IEEE Trans. Pattern Anal. Mach. Intell.}
  \bibinfo{volume}{35} (\bibinfo{year}{2013}) \bibinfo{pages}{1958--1971}.
\bibitem[{Lake et~al.(2013)Lake, Salakhutdinov, and Tenenbaum}]{LakSalTen13}
\bibinfo{author}{B.~M. Lake}, \bibinfo{author}{R.~Salakhutdinov},
  \bibinfo{author}{J.~B. Tenenbaum},
\newblock \bibinfo{title}{One-shot learning by inverting a compositional causal
  process},
\newblock in: \bibinfo{booktitle}{NIPS'13}, pp. \bibinfo{pages}{2526--2534}.
\bibitem[{Beyer et~al.(2009)Beyer, Cimatti, Griggio, Keremoglu, and
  Sebastiani}]{beyercgks09}
\bibinfo{author}{D.~Beyer}, \bibinfo{author}{A.~Cimatti},
  \bibinfo{author}{A.~Griggio}, \bibinfo{author}{M.~E. Keremoglu},
  \bibinfo{author}{R.~Sebastiani},
\newblock \bibinfo{title}{Software model checking via large-block encoding},
\newblock in: \bibinfo{booktitle}{FMCAD}, \bibinfo{publisher}{IEEE},
  \bibinfo{year}{2009}, pp. \bibinfo{pages}{25--32}.
\bibitem[{Franzen et~al.(2010)Franzen, Cimatti, Nadel, Sebastiani, and
  Shalev}]{fcnss_fmcad10}
\bibinfo{author}{A.~Franzen}, \bibinfo{author}{A.~Cimatti},
  \bibinfo{author}{A.~Nadel}, \bibinfo{author}{R.~Sebastiani},
  \bibinfo{author}{J.~Shalev},
\newblock \bibinfo{title}{{Applying SMT in Symbolic Execution of Microcode}},
\newblock in: \bibinfo{booktitle}{Proc. Int. Conference on Formal Methods in
  Computer Aided Design (FMCAD'10)}, \bibinfo{publisher}{IEEE},
  \bibinfo{year}{2010}.
\bibitem[{Fagerberg et~al.(2012)Fagerberg, Flamm, Merkle, and
  Peters}]{FagFlaMerPet12}
\bibinfo{author}{R.~Fagerberg}, \bibinfo{author}{C.~Flamm},
  \bibinfo{author}{D.~Merkle}, \bibinfo{author}{P.~Peters},
\newblock \bibinfo{title}{Exploring chemistry using smt},
\newblock in: \bibinfo{booktitle}{Proceedings of the 18th International
  Conference on Principles and Practice of Constraint Programming}, CP'12,
  \bibinfo{publisher}{Springer-Verlag}, \bibinfo{address}{Berlin, Heidelberg},
  \bibinfo{year}{2012}, pp. \bibinfo{pages}{900--915}.
\bibitem[{Yordanov et~al.(2013)Yordanov, Wintersteiger, Hamadi, and
  Kugler}]{YorWinHamKug13}
\bibinfo{author}{B.~Yordanov}, \bibinfo{author}{C.~M. Wintersteiger},
  \bibinfo{author}{Y.~Hamadi}, \bibinfo{author}{H.~Kugler},
\newblock \bibinfo{title}{Smt-based analysis of biological computation},
\newblock in: \bibinfo{booktitle}{NASA Formal Methods Symposium 2013}, volume
  \bibinfo{volume}{7871} of \textit{\bibinfo{series}{LNCS}},
  \bibinfo{publisher}{Springer Verlag}, \bibinfo{year}{2013}, pp.
  \bibinfo{pages}{78--92}.
\bibitem[{Yu and Joachims(2009)}]{YuJoa09}
\bibinfo{author}{C.-N.~J. Yu}, \bibinfo{author}{T.~Joachims},
\newblock \bibinfo{title}{Learning structural svms with latent variables},
\newblock in: \bibinfo{booktitle}{Proceedings of the 26th Annual International
  Conference on Machine Learning}, ICML '09, \bibinfo{publisher}{ACM},
  \bibinfo{address}{New York, NY, USA}, \bibinfo{year}{2009}, pp.
  \bibinfo{pages}{1169--1176}.
\bibitem[{Fr{\"{a}}nzle et~al.(2007)Fr{\"{a}}nzle, Herde, Teige, Ratschan, and
  Schubert}]{FranzleHTRS07}
\bibinfo{author}{M.~Fr{\"{a}}nzle}, \bibinfo{author}{C.~Herde},
  \bibinfo{author}{T.~Teige}, \bibinfo{author}{S.~Ratschan},
  \bibinfo{author}{T.~Schubert},
\newblock \bibinfo{title}{Efficient solving of large non-linear arithmetic
  constraint systems with complex boolean structure},
\newblock \bibinfo{journal}{{JSAT}} \bibinfo{volume}{1} (\bibinfo{year}{2007})
  \bibinfo{pages}{209--236}.
\bibitem[{Jovanovic and de~Moura(2012)}]{JovanovicM12}
\bibinfo{author}{D.~Jovanovic}, \bibinfo{author}{L.~M. de~Moura},
\newblock \bibinfo{title}{Solving non-linear arithmetic},
\newblock in: \bibinfo{editor}{B.~Gramlich}, \bibinfo{editor}{D.~Miller},
  \bibinfo{editor}{U.~Sattler} (Eds.), \bibinfo{booktitle}{IJCAR}, volume
  \bibinfo{volume}{7364} of \textit{\bibinfo{series}{Lecture Notes in Computer
  Science}}, \bibinfo{publisher}{Springer}, \bibinfo{year}{2012}, pp.
  \bibinfo{pages}{339--354}.
\bibitem[{Chavira and Darwiche(2008)}]{ChaDar08}
\bibinfo{author}{M.~Chavira}, \bibinfo{author}{A.~Darwiche},
\newblock \bibinfo{title}{On probabilistic inference by weighted model
  counting},
\newblock \bibinfo{journal}{Artif. Intell.} \bibinfo{volume}{172}
  (\bibinfo{year}{2008}) \bibinfo{pages}{772--799}.

\end{thebibliography}


\end{document}